\documentclass[11pt]{article}
\usepackage[final]{acl}
\usepackage{times}
\usepackage{latexsym}
\usepackage[T1]{fontenc}
\usepackage[utf8]{inputenc}
\usepackage{microtype}
\usepackage{inconsolata}
\usepackage{graphicx}
\usepackage{booktabs}
\usepackage{amsmath}
\usepackage{amssymb}
\usepackage{xcolor}
\usepackage{colortbl}
\usepackage{array}
\usepackage{tikz}
\usetikzlibrary{arrows.meta}
\usepackage[most]{tcolorbox}
\usepackage{multicol}
\usepackage{enumitem}
\newtcolorbox{casebox}[1]{
  colback=gray!3,
  colframe=gray!40,
  fonttitle=\bfseries\small,
  title=#1,
  boxrule=0.4pt,
  breakable,
  top=3mm,
  bottom=3mm,
  left=3mm,
  right=3mm,
  arc=1mm,
  before skip=10pt plus 2pt,
  after skip=10pt plus 2pt
}

\title{NovGauge: A Fine-Grained Benchmark for Diagnosing LLMs' Capability in Paper Novelty Assessment}

\author{
  \textbf{Guoqiang Zhang\textsuperscript{1,*,\ensuremath{\dagger}}},
  \textbf{Kexin Tan\textsuperscript{1,2,*}},
  \textbf{Ming Zhang\textsuperscript{1,*}},
  \textbf{Li Ju\textsuperscript{3}},
  \textbf{Wenqing Jing\textsuperscript{1}}, \\
  \textbf{Zhonghan Yue\textsuperscript{1}},
  \textbf{Jiayi Chen\textsuperscript{1}},
  \textbf{Shiqiang Wu\textsuperscript{1}},
  \textbf{Shaofan Liu\textsuperscript{1}},
  \textbf{Yue Zhang\textsuperscript{4}}, \\
  \textbf{Yuankai Ying\textsuperscript{1}},
  \textbf{Yang Shi\textsuperscript{3}},
  \textbf{Tao Gui\textsuperscript{1}},
  \textbf{Qi Zhang\textsuperscript{1,\ensuremath{\dagger}}},
  \textbf{Xuanjing Huang\textsuperscript{1}} \\[0.5em]
  \textsuperscript{1}College of Computer Science and Artificial Intelligence, Fudan University, Shanghai, China \\
  \textsuperscript{2}Graduate School of Engineering,
  The University of Tokyo, Tokyo, Japan \\
  \textsuperscript{3}AtomInnoLab 
  \textsuperscript{4}Atom Infinite Pte. Ltd., Singapore \\[0.3em]
  \texttt{zitago121@gmail.com}\quad\texttt{qz@fudan.edu.cn}
}

\let\novgaugeauthorblock\outauthor
\renewcommand{\outauthor}{%
  \begin{tabular}[t]{c}
    \novgaugeauthorblock\\[0.5em]
    {\normalfont\small
      \href{https://github.com/Zita-Go/NovGauge}{%
        \raisebox{-0.15em}{\includegraphics[height=1.1em]{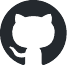}}\,
        \nolinkurl{https://github.com/Zita-Go/NovGauge}}\quad
      \href{https://huggingface.co/datasets/ZitaGo/NovGauge}{%
        \raisebox{-0.15em}{\includegraphics[height=1.1em]{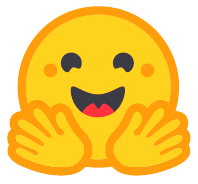}}\,
        \nolinkurl{https://huggingface.co/datasets/ZitaGo/NovGauge}}}
  \end{tabular}%
}

\begin{document}
\maketitle

\begin{abstract}
Large language models (LLMs) are increasingly used in peer review at major AI
conferences, yet novelty remains a persistent weak point.
Existing benchmarks assess novelty as a single holistic score, making it
difficult to diagnose which dimension a model misjudges or whether its evidence
is faithful.
We present \textsc{NovGauge}, a human-anchored benchmark for fine-grained
novelty assessment diagnosis.
The benchmark contains 619 paper pairs and 50 multi-paper sets, drawn from two expert sources: ICLR reviewer overlap
claims and survey co-citations.
Instances are independently labeled along three dimensions:
\emph{task}, \emph{problem}, and \emph{method}, capturing application goals,
technical challenges, and solution approaches.
We propose a cascading diagnostic pipeline that verifies per-dimension
correctness, evidence grounding, and logical support.
Evaluation of 18 LLMs shows hallucination rates ranging from 0\% to 39\%
across dimensions, and among non-hallucinated correct-positive judgments, over
70\% cite evidence fails to logically support the stated reason.
The best-performing model, GPT-5.5, achieves 43--72\% Verified F1 across
dimensions, while most models retain less than half of their raw F1 after
faithfulness verification.
These results suggest that current LLMs remain far from reliable scientific
novelty assessment, particularly when correctness is conditioned on faithful
evidence grounding.
\end{abstract}

\section{Introduction}
\label{sec:intro}

\begin{figure}[!t]
  \centering
  \includegraphics[width=\columnwidth,trim=160pt 6pt 110pt 4pt,clip]{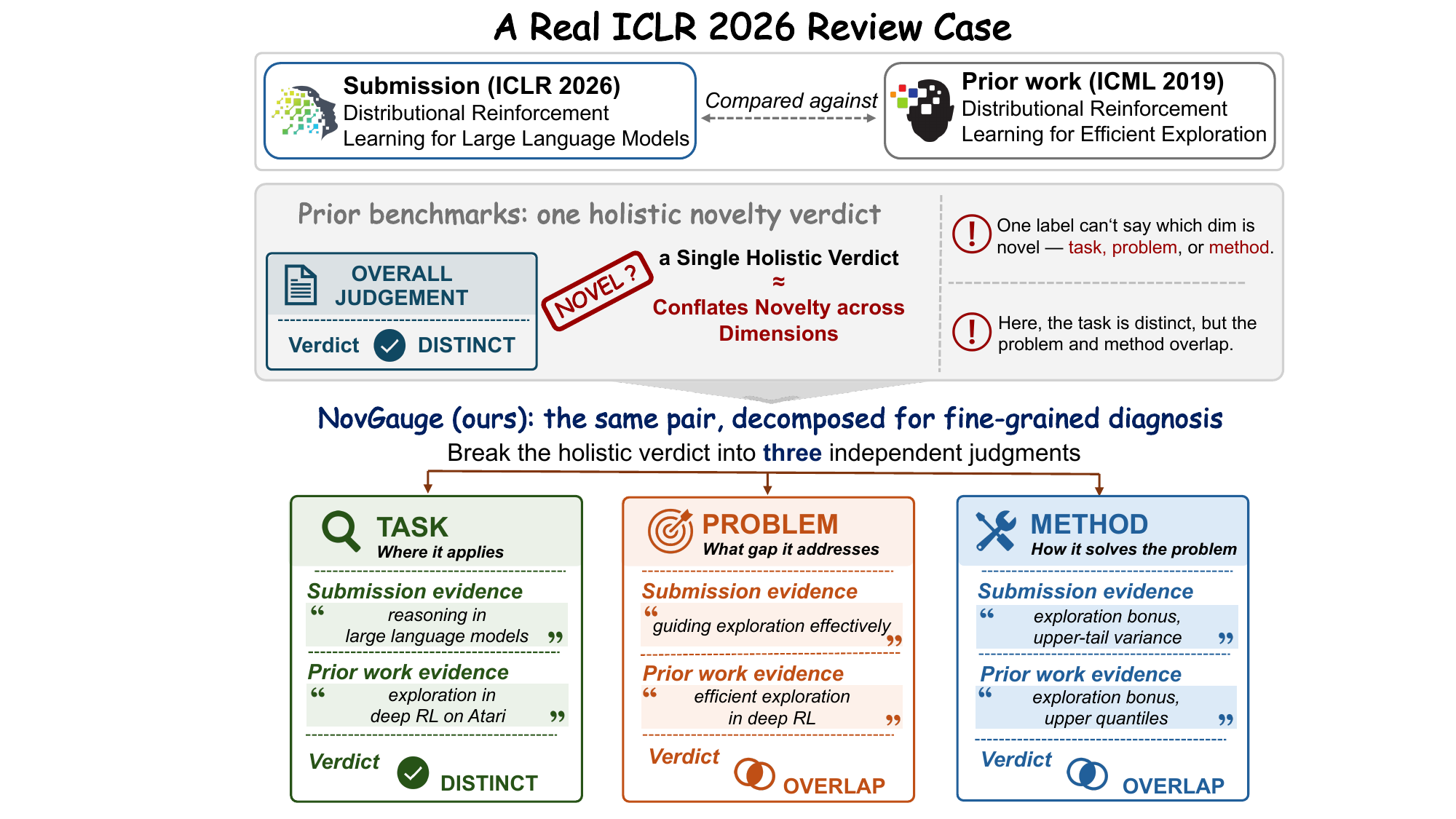}
  \caption{\textbf{One label, three dimensions.}
  An ICLR~2026 submission compared against flagged prior work.
  \textsc{NovGauge} evaluates each dimension independently, revealing that
  this pair shares its method while targeting a distinct task.}
  \label{fig:teaser}
\end{figure}

\begin{figure*}[t]
  \centering
  \includegraphics[width=0.95\textwidth,trim=0pt 0pt 0pt 20pt,clip]{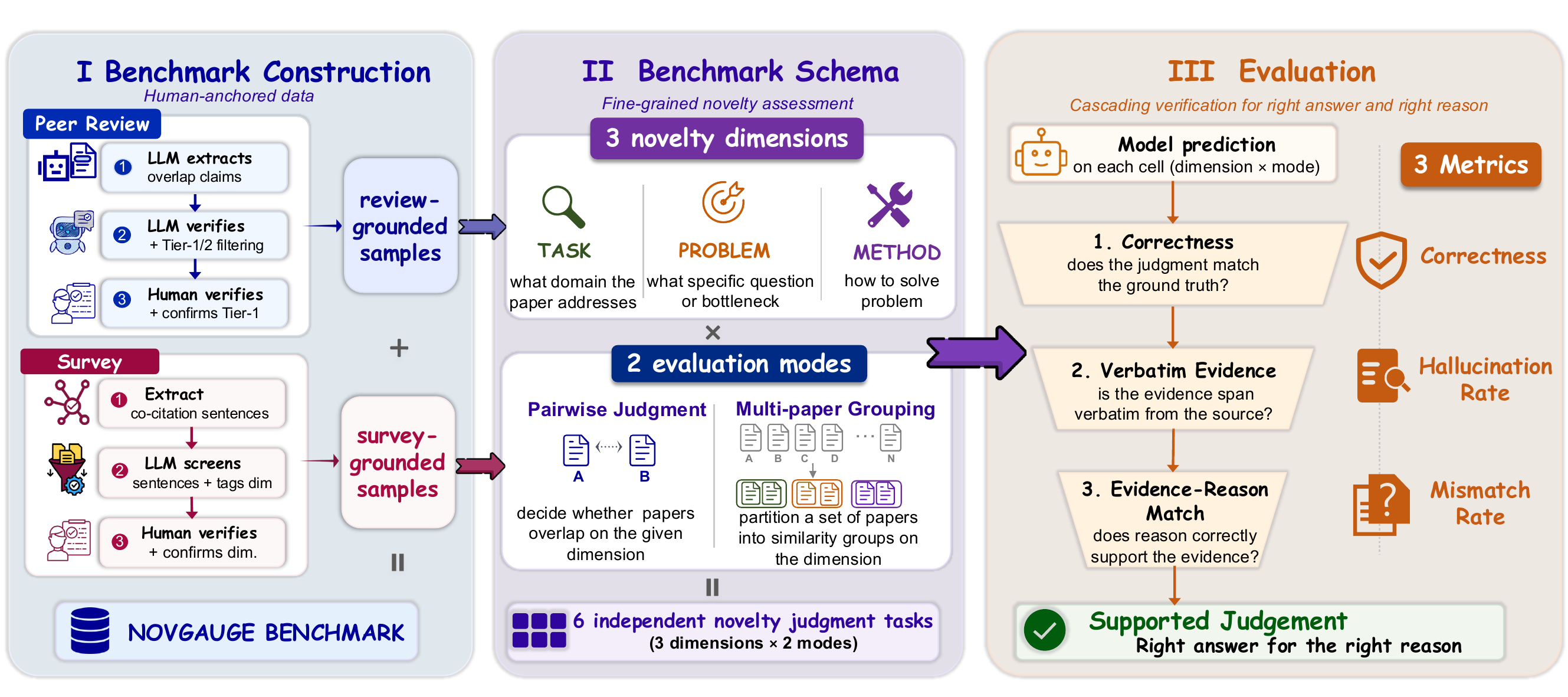}
  \caption{\textbf{Overview of the \textsc{NovGauge} framework.}
    Two human-anchored construction pipelines produce pairs and multi-paper
    sets with task, problem, and method labels. Model outputs pass through a cascading 
    evaluation funnel that checks correctness, evidence hallucination, and 
    reasoning mismatch, yielding a per-dimension diagnosis.}
  \label{fig:pipeline}
\end{figure*}

Large language models already shape peer review at scale through two channels.
Reviewers increasingly rely on general-purpose LLMs to draft reviews, with
audits estimating that up to 20\% of text at major AI venues is now
LLM-generated~\citep{liang2024monitoring,shen2026detecting}.
Meanwhile, dedicated automated review systems continue to
emerge~\citep{zhou-etal-2024-llm,darcy2024margmultiagentreviewgeneration,zhu-etal-2025-deepreview}, with some
already approaching human-level agreement on acceptance
decisions~\citep{weng2025cycleresearcher}.
However, the reliability of LLM-generated reviews remains questionable~\cite{wu2026aigoodpeerreviewer,shin2025mindblindspotsfocuslevel,du2024llms}.
Recent evaluations show that LLM reviews correlate weakly with human judgments
and can be trivially gamed~\citep{baumann2026stop}.
This concern is especially important for novelty assessment: large-scale
analysis of ICLR 2024--2025 identifies lack of novelty as one of the most
frequently cited weaknesses in rejected papers~\citep{kargaran2025iclrinsights}.

Despite its importance, novelty remains a particularly challenging criterion for LLM reviewers.
Studies show that LLMs systematically underweight novelty in their
reviews~\citep{shin-etal-2025-mind} and that their novelty ratings even
correlate negatively with a paper's eventual impact~\citep{jiang2026hindsight}.
These results confirm that LLM novelty assessment is unreliable, but a single
accuracy number on a holistic label reveals only that a model underperforms,
not which aspect of novelty it misjudges.
This leaves the operational review question unanswered: given a submission and
a flagged prior work, which novelty dimension overlaps, and is the cited
evidence faithful?

Reliable diagnosis demands two capabilities.
The first is \emph{per-dimension} evaluation.
Novelty is not monolithic: scientometric studies decompose it into research
question and method~\citep{luo2022combination,shen2026ideation}, and a paper may
overlap with prior work on one dimension while remaining novel on others
(Figure~\ref{fig:teaser}).
Current benchmarks, however, either let dimensions vary across
instances~\citep{moussa2025scholareval} or treat novelty as one undifferentiated
score among several review
criteria~\citep{qiao2026innoeval,schopf2026rinobench,lin2025schnovel,wu2026novbench}.
The second is \emph{faithfulness} verification.
Even when a model's judgment is correct, the reasoning behind it may rest on
fabricated evidence or logical gaps. No existing benchmark systematically
diagnoses whether failures stem from judgment errors, fabricated evidence, or
logical gaps~\citep{schopf2026rinobench}.

We introduce \textsc{NovGauge}, a benchmark designed to enable fine-grained
diagnosis of novelty judgment.
Every instance is independently labeled for task, problem, and method,
providing the fixed per-dimension ground truth.
Labels are anchored in two complementary human sources, ICLR reviewer overlap
claims and annotator-verified survey co-citations.
Beyond pairwise comparison, the benchmark further includes 50 multi-paper
grouping sets.
Rather than stopping at binary correctness, the evaluation protocol applies a
\emph{cascading} diagnostic pipeline that checks correctness, evidence 
hallucination, and reasoning mismatch.
Only predictions that survive all three stages count toward \textbf{Verified F1},
ensuring the model reached \emph{the right answer for the right reason}.

Our contributions are:
\begin{itemize}
\item A \textbf{benchmark of 619 pairs and 50 multi-paper sets} with 
fixed three-dimension ground truth, anchored in
ICLR reviewer statements and annotator-verified survey co-citations.
\item A \textbf{cascading evaluation protocol} that diagnoses novelty judgment
per dimension and audits evidence faithfulness through hallucination and
mismatch checks.
\item A \textbf{systematic evaluation of 18 LLMs} revealing that even correct
judgments are frequently backed by fabricated or logically disconnected
evidence, with failure patterns varying sharply across dimensions.
\end{itemize}

Figure~\ref{fig:pipeline} summarizes the benchmark construction and evaluation
flow.

\section{Related Work}
\label{sec:related}

\subsection{Automated Paper Review}

LLM-based peer review has advanced from single-model critique
generation~\citep{zhou-etal-2024-llm} to multi-agent
architectures~\citep{darcy2024margmultiagentreviewgeneration,yu2024sea,bougie2024generative} and fine-tuned models trained on
structured reasoning chains~\citep{zhu-etal-2025-deepreview,weng2025cycleresearcher,taechoyotin2025remor}.
However, systematic analyses reveal that LLMs over-attend to technical
validity while underweighting novelty~\citep{shin-etal-2025-mind},
assign inflated scores to LLM-generated
papers~\citep{li2025llmrevaltrustllmreviewers}, and align with human judgments only on
lower-quality submissions~\citep{doi:10.1056/AIoa2400196}.
\citet{li2026ratingcomprehensiveevaluationbenchmark} further show that aligning critique focus
with human experts is a prerequisite for reliable scoring.
These findings motivate benchmarks that diagnose novelty judgment
specifically.

\subsection{Novelty and Similarity Benchmarks}

Table~\ref{tab:benchmark_comparison} situates our work within a progression
from document-level similarity to fine-grained novelty assessment.
SciDocs~\citep{cohan-etal-2020-specter} and SciRepEval~\citep{singh2023scirepeval}
evaluate retrieval relevance from citation-graph labels, measuring similarity
holistically without distinguishing which dimensions overlap.

Recent novelty benchmarks move beyond retrieval to assess whether a paper or
idea advances the state of the art.
NovBench~\citep{wu2026novbench} frames novelty as text generation from a
single paper's introductory claim, while
literature-grounded systems assess manuscript novelty against existing
work~\citep{mostafa2026whatisnovel,zhang2026opennovelty}.
Idea-level benchmarks evaluate research ideas against retrieved related works:
RINoBench~\citep{schopf2026rinobench} and InnoEval~\citep{qiao2026innoeval}
assign novelty scores, while ScholarEval~\citep{moussa2025scholareval} and
the Idea Novelty Checker~\citep{shahid2025novelty} dynamically identify
contribution dimensions for each idea.
Pairwise approaches take designated paper pairs as input but target review
quality assessment~\citep{zhang2025pairwise,zheng2026cnpe} or relative
novelty ranking~\citep{lin2025schnovel} rather than overlap detection on
fixed dimensions.

\begin{table*}[t]
\centering
\small
\setlength{\tabcolsep}{4pt}
\begin{tabular}{l c c c c c r}
\toprule
\textbf{Benchmark} & \textbf{Task} & \textbf{Pairwise} & \textbf{Grouping} & \textbf{Expert} & \textbf{\#Dim.} & \textbf{Scale} \\
\midrule
SciDocs \citep{cohan-etal-2020-specter}                               & Retrieval       & \checkmark  & ---         & ---         & 1     & 25K papers \\
SciRepEval \citep{singh2023scirepeval}                         & Retrieval       & \checkmark  & ---         & ---         & 1     & 60K papers \\
NovBench \citep{wu2026novbench}                                & Generation      & ---         & ---         & \checkmark  & 1     & 1.7K pairs \\
RINoBench \citep{schopf2026rinobench}                          & Scoring         & ---         & ---         & \checkmark  & 1     & 1.4K ideas \\
ScholarEval \citep{moussa2025scholareval}                      & Scoring + RAG   & ---         & ---         & \checkmark  & var.  & 117 ideas \\
Idea Novelty Checker \citep{shahid2025novelty}                 & Scoring + RAG   & ---         & ---         & \checkmark  & 1     & 67 ideas \\
InnoEval \citep{qiao2026innoeval}                              & Scoring         & ---         & ---         & \checkmark  & 1     & 217 ideas \\
SchNovel \citep{lin2025schnovel}                               & Ranking         & \checkmark  & ---         & ---         & 1     & 15K pairs \\
\midrule
\rowcolor{gray!15}
\textbf{Ours (pairwise)} & \textbf{Classification} & \checkmark & --- & \checkmark & \textbf{3} & \textbf{619 pairs} \\
\rowcolor{gray!15}
\textbf{Ours (grouping)} & \textbf{Grouping} & --- & \textbf{\checkmark} & \checkmark & \textbf{3} & \textbf{50 sets} \\
\bottomrule
\end{tabular}
\caption{\textbf{Comparison with existing novelty and similarity benchmarks.}
\textbf{\#Dim.} denotes the number of orthogonal dimensions along which novelty overlap is independently judged. These dimensions include task, problem, and method, rather than general evaluation rubrics such as relevance or clarity.
``var.'' denotes dimensions that vary across instances.}
\label{tab:benchmark_comparison}
\end{table*}

\textsc{NovGauge} addresses these gaps by providing per-instance ground truth 
under a fixed three-dimension taxonomy, enabling systematic diagnosis of where 
and why LLM novelty judgments fail.

\section{Benchmark Construction}
\label{sec:dataset}


\begin{table}[t]
\centering\footnotesize
\setlength{\tabcolsep}{3pt}
\begin{tabular}{@{}lrl@{}}
\toprule
\textbf{Source} & \textbf{Size} & \textbf{Main Dim.} \\
\midrule
\rowcolor{gray!10}
\multicolumn{3}{@{}l}{\textit{Pairwise}} \\
ICLR review positives        & 236 & method and problem \\
Survey co-citation positives & 227 & task \\
Survey cross-group negatives & 156 & method and task \\
\midrule
\rowcolor{gray!10}
\multicolumn{3}{@{}l}{\textit{Multi-paper grouping}} \\
Survey co-citation sets      & 50  & task, problem, method \\
\bottomrule
\end{tabular}
\caption{Final benchmark composition. ``Main Dim.''\ denotes the dominant dimension in each subset.}
\label{tab:dataset}
\end{table}

\subsection{Design Principles}
\label{sec:design}

\textsc{NovGauge} addresses the limitations of existing benchmarks through three 
core design principles.

\paragraph{Principle 1: Per-dimension decomposition.}
Novelty is not monolithic.
We decompose it into three orthogonal dimensions and evaluate each one independently.
\begin{itemize}
  \item \textbf{Task}: the application domain (e.g., \emph{image classification}
        vs.\ \emph{machine translation}).
  \item \textbf{Problem}: the core research challenge (e.g., \emph{distribution
        shift} vs.\ \emph{catastrophic forgetting}).
  \item \textbf{Method}: the technical approach (e.g., \emph{contrastive learning}
        vs.\ \emph{supervised fine-tuning}).
\end{itemize}
This follows the \emph{what}, \emph{why}, and \emph{how} of a research contribution
and extends prior taxonomies that separate research question and method
\citep{luo2022combination, shen2026ideation}. We treat task as a separate axis because
papers may share a problem or method while targeting different domains. Other
contribution types are represented within these dimensions rather than treated as
separate axes.
Figure~\ref{fig:teaser} illustrates partial overlap across dimensions.
For this reason, every instance carries fixed per-dimension ground truth, enabling us
to identify which dimension a model misjudges.

\paragraph{Principle 2: Expert-sourced labels.}
Every label is anchored in explicit expert evidence.
Positive labels originate from two complementary sources: ICLR reviewer 
overlap claims from public reviews, and survey co-citation groupings verified 
by three annotators.
Negatives are constructed through cross-group sampling, where pairs are drawn from 
papers a survey author placed into \emph{distinct} similarity groups within the 
same sentence.

\paragraph{Principle 3: Evidence-grounded evaluation.}
Correct judgments are insufficient if the reasoning behind them is unreliable.
We therefore require models to ground every positive judgment in verbatim evidence 
spans and evaluate outputs through the three-stage cascade described in 
Section~\ref{sec:eval_protocol}.


\subsection{Data Construction}
\label{sec:data_construction}

\textsc{NovGauge} is constructed from two complementary sources 
(Table~\ref{tab:dataset}), both verified by three PhD-level annotators.
ICLR-grounded pairs predominantly capture method and problem overlap, while 
survey-derived pairs are dominated by task-level similarity.
The ICLR source contributes positives only.
The survey source supplies both positives and dimension-specific negatives.

\textbf{Source 1: ICLR Review-Grounded Pairs.}
We process 42{,}682 ICLR 2023--2026 OpenReview forums through a two-stage LLM 
pipeline that extracts and verifies novelty-overlap claims from official reviews 
and author rebuttals.
Three PhD-level annotators adjudicate every retained candidate, verifying prior-work 
identity, dimension labels, and reviewer evidence.
The released test set retains only high-confidence pairs with convergent evidence 
from multiple reviewers and author concession, yielding \textbf{236 pairs}.
Detailed annotation procedures are in Appendix~\ref{app:iclr_construction}.

\textbf{Source 2: Survey Co-citation Groups.}
We extract similarity groupings from 459 CS arXiv surveys (2023--2024).
A rule-based extractor identifies 1{,}130 co-citation sentences with explicit 
similarity cues, and an LLM proposes per-dimension groupings.
The same annotators verify each grouping, yielding \textbf{118 verified 
sentences}.
We enumerate within-group pairs as positives (\textbf{227 pairs}) and cross-group 
pairs as negatives (\textbf{156 pairs}).
The survey source also contributes \textbf{85 multi-paper grouping instances} 
derived from 50 multi-paper sets (Section~\ref{sec:task}).
Detailed pipeline is in Appendix~\ref{app:survey_construction}.

\textbf{Final composition and quality control.}
Table~\ref{tab:dataset} summarizes the four subsets that make up the benchmark.
At the paper-pair level, the pairwise subset contains 619 records, of which 463
are labeled positive and 156 are labeled negative. Because each available
dimension is evaluated independently, these records yield 875
dimension-specific evaluation instances, of which 649 carry positive labels and
226 carry negative labels.
To verify label consistency, we audited a random sample of 30 items stratified
by dimension. The same annotators re-examined the labels independently, yielding
93.3\% agreement with the original annotations. Table~\ref{tab:dim_agreement} reports 
the per-dimension breakdown of this audit and the main causes of disagreement.


\subsection{Evaluation Formulation}
\label{sec:task}

We define two complementary evaluations, both scored independently per dimension.
The first tests pairwise comparison, where a model decides whether a submission
and a specific prior work overlap on the target dimension.
The second requires synthesizing evidence across multiple documents, a more realistic setting that mirrors how reviewers survey related work.

\paragraph{Pairwise Novelty Judgment.}
\textbf{Input}: a pair of papers $(A, B)$ and a target dimension 
$d \in \{\text{task}, \text{problem}, \text{method}\}$, with each paper rendered 
at one of three content granularities (abstract, abstract + introduction, or 
full paper).
A separate model call, with a dimension-specific prompt, is issued for each
available labeled dimension. A paper-pair record with labels on multiple
dimensions therefore contributes one dimension-specific evaluation instance per
label rather than one holistic instance.

\textbf{Output}: a structured JSON record judging similarity on \emph{$d$} 
and grounding it in verbatim evidence:
\begin{quote}\small\ttfamily
\{ is\_similar: bool,\\
\hspace*{1em}evidence\_a: <verbatim span from A>,\\
\hspace*{1em}evidence\_b: <verbatim span from B>,\\
\hspace*{1em}reason: <$\leq$2-sentence justification> \}
\end{quote}
Predictions are evaluated against the ground-truth labels in 
Table~\ref{tab:dataset}.

\paragraph{Multi-paper Similarity Grouping.}
Given a \emph{set} of related papers, the model must identify which subsets share a given dimension.

\textbf{Input}: a set of $N$ papers ($3 \leq N \leq 10$), together with a target dimension $d$.
The $N$ papers are not all mutually similar. Only certain subsets share the dimension.

\textbf{Output}: the model must partition the $N$ papers into similarity subgroups 
for dimension $d$:
\begin{quote}\small\ttfamily
\{ groups: [\\
\hspace*{1em}\{ paper\_indices: [\ldots],\\
\hspace*{2em}evidence: \{\ldots\}, reason: \ldots \}, \ldots ] \}
\end{quote}
Each output subgroup is a maximal set of papers judged mutually similar on $d$.
Each input set is evaluated on every dimension for which a ground-truth 
label exists, yielding 85 grouping instances across the 50 input sets 
(44 task, 13 problem, 28 method).


\subsection{Cascading Evaluation Protocol}
\label{sec:eval_protocol}

A model may produce a correct label with fabricate supporting evidence, or cite real 
evidence that does not logically support its conclusion.
To separate grounded reasoning from surface-level correctness, we evaluate outputs along 
three cascading stages (Figure~\ref{fig:pipeline}, right panel), each filtering the 
outputs of the previous stage.

\textbf{Stage 1: Correctness.}
For pairwise judgment, we report per-dimension Accuracy and F1 for classifying
whether a pair is similar.
For multi-paper grouping, each of the 85 (set, dimension) instances yields a predicted 
partition, scored against the ground-truth partition by converting both into their 
sets of within-set paper pairs and computing pairwise Precision, Recall, and F1 
($\text{TP} = |\text{pred pairs} \cap \text{GT pairs}|$).
These per-instance F1 scores are then macro-averaged within each dimension.
For both evaluations, every reported metric is the mean across $n{=}3$ inference samples.

\textbf{Stage 2: Hallucination Rate.}
Every positive output must cite verbatim evidence from the source papers. 
For pairwise predictions, this means one span per paper. For grouping 
predictions, one span per member in each predicted subgroup.
Each cited span is verified against the source paper's text via substring 
matching.
A span is classified \emph{hallucinated} if verification fails.
The Hallucination Rate is the fraction of positive outputs (pairs or subgroups) in 
which at least one cited span is hallucinated.
Details are in Appendix~\ref{app:evidence_verification}.

\textbf{Stage 3: Mismatch Rate.}
Among the \emph{non-hallucinated} positive outputs, an external LLM judge decides 
whether the stated reason is logically entailed by the cited evidence.
The judge model and prompt are detailed in Section~\ref{sec:experiments} and
Appendix~\ref{app:prompts}.

\textbf{Verified F1.}
The three stages compose into a single summary metric: \textbf{Verified F1}.
Only true positives that pass all three stages count in the numerator, while 
hallucinated or mismatched predictions remain in the denominator.
For multi-paper grouping, subgroups are first classified by their hallucination 
and mismatch status, then converted to pairs for F1 computation.
The gap between raw F1 and Verified F1 quantifies the \emph{faithfulness cliff}, 
the fraction of seemingly correct judgments undermined by fabricated or unsupported 
evidence.
Formal definitions are in Appendix~\ref{app:metrics}.

\section{Experimental Setup}
\label{sec:experiments}

We evaluate \textsc{NovGauge} through three experiments. The first tests
pairwise novelty judgment, the main task. The second varies how much paper
content is shown to the model. The third tests whether models can group several
papers that overlap along the same novelty dimension. All experiments use the
cascading protocol in \S\ref{sec:eval_protocol}.

\subsection{Pairwise Judgment}
\label{sec:exp:pairwise}

Pairwise judgment asks a model to decide whether two papers overlap in task,
problem, or method. We evaluate 18 LLMs spanning frontier systems, mid-scale
open-weight models, and specialised peer-review models. Each model produces a
binary label, a short reason, and evidence spans from the papers. We report raw
F1, Hallucination Rate, Mismatch Rate, and Verified F1. The full model list and
system settings appear in Appendix~\ref{app:models}.

\subsection{Granularity Ablation}
\label{sec:exp:granularity}

This experiment tests whether more paper context improves novelty judgment. We
run the same pairwise task under three input settings: abstract only,
abstract plus introduction, and full paper. The corresponding per-paper content caps are 1k, 4k, and 16k tokens. The full-paper setting is the default for the main pairwise results.

\subsection{Multi-Paper Grouping}
\label{sec:exp:multipaper}

Multi-paper grouping asks a model to recover clusters of papers that share the
same task, problem, or method overlap. This setting is harder than pairwise
judgment because the model must compare multiple papers and provide collective
evidence for each predicted group. We evaluate the 12 frontier models on 85
grouping instances, using full paper inputs.

\begin{table*}[t]
\centering
\footnotesize
\setlength{\tabcolsep}{2.5pt}
\begin{tabular}{l ccc>{\columncolor{blue!8}}c ccc>{\columncolor{blue!8}}c ccc>{\columncolor{blue!8}}c}
\toprule
& \multicolumn{4}{c}{\textbf{Task}}
& \multicolumn{4}{c}{\textbf{Problem}}
& \multicolumn{4}{c}{\textbf{Method}} \\
\cmidrule(lr){2-5}\cmidrule(lr){6-9}\cmidrule(lr){10-13}
\textbf{Model}
& F1$\uparrow$ & Hal$\downarrow$ & Mis$\downarrow$ & VF1$\uparrow$
& F1$\uparrow$ & Hal$\downarrow$ & Mis$\downarrow$ & VF1$\uparrow$
& F1$\uparrow$ & Hal$\downarrow$ & Mis$\downarrow$ & VF1$\uparrow$ \\
\midrule
\rowcolor{gray!10}
\multicolumn{13}{l}{\textsc{Frontier}} \\
GPT-5.5$^\ddagger$
& 84.8 & 0.0           & \textbf{15.7} & \textbf{71.5}
& 69.4 & \textbf{0.0}  & \textbf{12.1} & \textbf{61.0}
& 78.7 & \textbf{0.0}  & \textbf{45.3} & \textbf{43.1} \\
GPT-5.4-mini$^\ddagger$
& 68.1 & 0.0 & 23.4 & 52.1
& 56.7 & 1.0 & 33.5 & 37.3
& 60.8 & 0.3 & 51.7 & 29.2 \\
Claude Opus 4.7
& 77.7 & 0.0 & 76.6 & 18.2
& 66.2 & \textbf{0.0} & 84.2 & 10.5
& 72.2 & 5.8 & 86.7 & 9.1 \\
Claude Sonnet 4.6
& 84.3 & 0.2 & 78.1 & 18.4
& 66.1 & 1.7 & 79.4 & 13.4
& 70.4 & 23.2 & 84.0 & 8.6 \\
Gemini 3.1 Pro
& 84.5 & 0.0 & 60.3 & 33.5
& 77.1 & 1.9 & 67.3 & 24.7
& 78.0 & 3.6 & 73.5 & 19.9 \\
Gemini 3.1 Flash-Lite
& 85.8 & 1.1 & 73.1 & 22.8
& \textbf{89.2} & 5.8 & 72.8 & 22.9
& \textbf{81.9} & 8.9 & 79.2 & 15.5 \\
Kimi-K2.6
& \textbf{87.6} & 0.0 & 77.3 & 19.9
& 76.5 & 2.0 & 65.8 & 25.7
& 64.6 & 6.6 & 78.6 & 12.9 \\
GLM-5.1
& 71.2 & 0.0 & 66.3 & 24.0
& 69.3 & 1.6 & 62.8 & 25.3
& 57.4 & 27.2 & 77.4 & 9.4 \\
Qwen3.6-Plus
& 83.0 & 0.0 & 73.5 & 22.0
& 74.9 & 2.4 & 83.2 & 12.3
& 63.0 & 4.0 & 86.4 & 8.2 \\
DeepSeek-V4-Pro
& 61.9 & 0.0 & 67.8 & 20.0
& 60.3 & 0.9 & 81.0 & 11.4
& 51.2 & 24.9 & 82.7 & 6.7 \\
DeepSeek-V4-Flash
& 66.3 & 0.0 & 72.7 & 18.1
& 60.0 & \textbf{0.0} & 93.8 & 3.7
& 61.3 & 17.3 & 89.2 & 5.4 \\
MiniMax-M2.7
& 73.2 & 0.3 & 86.4 & 9.9
& 73.3 & 0.7 & 93.9 & 4.4
& 64.3 & 14.0 & 89.9 & 5.5 \\
\midrule
\rowcolor{gray!10}
\multicolumn{13}{l}{\textsc{Mid-scale open-weight}} \\
Qwen3-8B
& 70.3 & 3.2 & 85.8 & 9.6
& 53.4 & 6.2 & 92.2 & 3.9
& 48.1 & 10.9 & 90.0 & 4.3 \\
Qwen3-32B
& 61.0 & 2.0 & 75.7 & 14.5
& 56.6 & 2.1 & 70.7 & 16.2
& 49.3 & 12.7 & 79.4 & 8.9 \\
Qwen3.5-27B
& 76.2 & 0.2 & 81.2 & 14.3
& 76.7 & 1.3 & 86.0 & 10.6
& 67.1 & 4.8 & 87.1 & 8.2 \\
Gemma4-31B-IT
& 73.7 & 0.0 & 52.6 & 34.9
& 66.2 & 0.8 & 51.7 & 31.7
& 64.1 & 4.5 & 68.7 & 19.2 \\
\midrule
\rowcolor{gray!10}
\multicolumn{13}{l}{\textsc{Specialised}} \\
CycleReviewer-8B
& 20.6 & 26.0 & 84.2 & 2.4
& 22.5 & 24.2 & 83.0 & 2.9
& 10.0 & 39.0 & 92.0 & 0.5 \\
DeepReviewer-14B
& 60.7 & 8.7 & 71.3 & 15.9
& 73.1 & 28.6 & 70.0 & 15.7
& 41.9 & 11.1 & 74.0 & 9.7 \\
\bottomrule
\end{tabular}
\caption{\textbf{Pairwise novelty judgment results} (full-paper content granularity).
\textbf{Hal} denotes Hallucination Rate, \textbf{Mis} denotes Mismatch Rate, and
\textbf{VF1} denotes Verified F1.
Definitions appear in \S\ref{sec:eval_protocol}.
All metrics reported as percentages.
\textbf{Bold} marks the best value per column when fewer than half the models tie.
$^\ddagger$GPT-family judged by Gemini-3.1-Pro.
Standard deviations appear in Appendix~\ref{app:detailed_results}.}
\label{tab:main_pairwise}
\end{table*}

\subsection{Implementation Details}
\label{sec:exp:setup}

Each input and dimension is decoded three times at temperature 0.6, and we
report mean scores with standard deviations in Appendix~\ref{app:detailed_results}.
All models use their maximum reasoning setting when available. The mismatch
stage uses an LLM judge at temperature 0.2 (prompt in Appendix~\ref{app:prompts}). GPT-family models are judged by
Gemini-3.1-Pro, while all other models are judged by GPT-5.2. A human validation
study on 100 non-hallucinated positive outputs finds 87.0\% agreement with the
judge, with Cohen's $\kappa=0.74$ (details in Appendix~\ref{app:judge_validation}).

\section{Experimental Results}
\label{sec:results}

\subsection{Main Results}
\label{sec:res:pairwise}

\begin{figure}[t]
\centering
\includegraphics[width=\columnwidth]{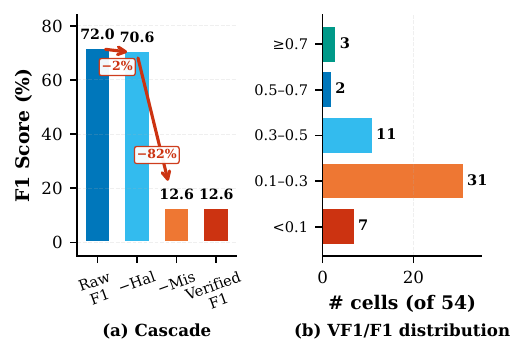}
\caption{Panel (a) shows the faithfulness cascade for Claude~Opus~4.7 on the
method dimension, where Raw F1 drops to low Verified F1 after hallucination and
mismatch checks.
Panel (b) shows the distribution of VF1/F1 ratios across the 54 model-dimension
cells of Table~\ref{tab:main_pairwise}.}
\label{fig:cascade}
\label{tab:cliff}
\end{figure}

Table~\ref{tab:main_pairwise} reports pairwise novelty judgment results for the
18 evaluated models across task, problem, and method. Raw F1 is often high, but
the cascading evaluation exposes a large gap between correct labels and faithful reasoning
(Figure~\ref{fig:cascade}a).
The average model retains only 26\% of its raw F1 after faithfulness filtering.

\paragraph{Faithfulness filtering exposes unsupported reasoning.}
\label{sec:analysis:cliff}
Only 5 of 54 model-dimension cells retain at least half of their raw F1 after
faithfulness filtering, as shown in Figure~\ref{fig:cascade}b. The main source of degradation is mismatch. 
The mismatch stage flags 72.5\% of non-hallucinated correct-positive predictions
as unsupported because models identify
similarity at the contribution level but cite evidence at the implementation
level. For example, Gemini~3.1~Flash-Lite
reaches 89.2\% raw F1 on problem but drops to 22.9\% Verified F1. GPT-5.5 is
the clear exception, with zero detected hallucinations and mismatch rates below
46\% across all dimensions.
Notably, CycleReviewer-8B and DeepReviewer-14B reach at
most 15.9\% Verified F1, suggesting that review-oriented fine-tuning does not
necessarily transfer to fine-grained, evidence-grounded novelty judgment.

\paragraph{Method-level novelty is the hardest to ground.}
The method dimension is the least faithful dimension for nearly all models.
Hallucination on method often exceeds task and problem by a wide margin. For
GLM-5.1, the method hallucination rate is 27.2\%, compared with 0.0\% on task.
Even GPT-5.5, the strongest model overall, retains 84\% of its raw F1 on task
but only 55\% on method, confirming that
method-level novelty is the most fragile setting for LLM-based review support.

\begin{figure}[t]
\centering
\includegraphics[width=\columnwidth]{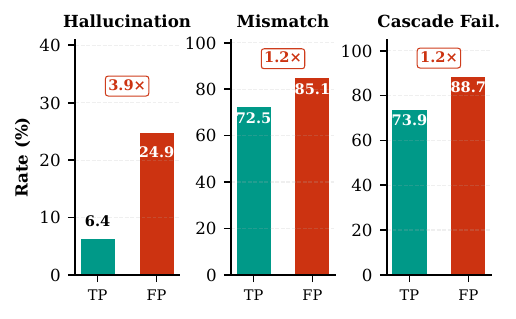}
\caption{Faithfulness on correct versus incorrect positive predictions.
Panel mean over 18 models.}
\label{fig:tp_fp_hallucination}
\end{figure}

\paragraph{Incorrect positive predictions fabricate more evidence.}
\label{sec:analysis:reflex}
Figure~\ref{fig:tp_fp_hallucination} compares faithfulness on correct and
incorrect positive predictions. Hallucination Rate rises from 6.4\% on true
positives to 24.9\% on false positives. Mismatch Rate also increases, from
72.5\% to 85.1\%. Only 11\% of false-positive predictions survive the full
cascade, compared with 26\% of true positives. This suggests that false
positives are not merely label errors, they often combine an unsound conclusion
with weaker evidence selection.

\begin{figure}[t]
\centering
\includegraphics[width=0.9\columnwidth]{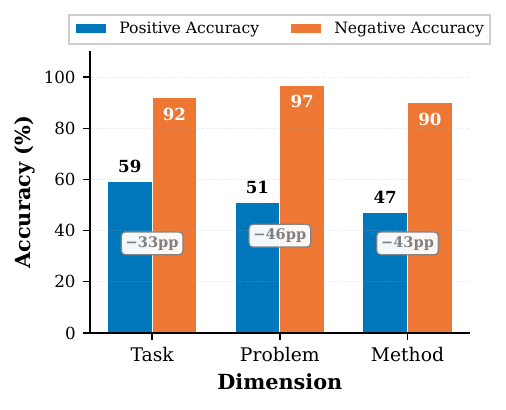}
\caption{Positive accuracy (correctly identifying similar pairs) versus negative
accuracy (correctly rejecting not-similar pairs) across three dimensions, averaged
over 18 models.}
\label{fig:pos_neg_acc}
\end{figure}

\paragraph{LLMs are conservative similarity judges.}
\label{sec:analysis:conservative}
Models achieve 90--97\% negative accuracy but only 47--59\% positive accuracy,
as shown in Figure~\ref{fig:pos_neg_acc}. The gap also holds within the survey source,
which contains both labels: positive accuracy is 44--59\%, compared with 90--97\% for
negatives (Table~\ref{tab:source_controlled}). Among the 61 pairs missed by all 18
LLMs, every case is a false negative.  In other words, LLMs are
far more likely to miss genuine similarity than to fabricate spurious overlap,
particularly when two papers differ in framing or implementation despite sharing
a core contribution.

\begin{table*}[t]
\centering\small
\setlength{\tabcolsep}{3pt}
\begin{tabular}{l ccc>{\columncolor{blue!8}}c ccc>{\columncolor{blue!8}}c ccc>{\columncolor{blue!8}}c}
\toprule
& \multicolumn{4}{c}{\textbf{Task}}
& \multicolumn{4}{c}{\textbf{Problem}}
& \multicolumn{4}{c}{\textbf{Method}} \\
\cmidrule(lr){2-5} \cmidrule(lr){6-9} \cmidrule(lr){10-13}
\textbf{Model}
& F1$\uparrow$ & Hal$\downarrow$ & Mis$\downarrow$ & VF1$\uparrow$
& F1$\uparrow$ & Hal$\downarrow$ & Mis$\downarrow$ & VF1$\uparrow$
& F1$\uparrow$ & Hal$\downarrow$ & Mis$\downarrow$ & VF1$\uparrow$ \\
\midrule
GPT-5.5$^\ddagger$
& 77.1 & 0.0           & \textbf{42.4} & \textbf{41.7}
& 62.6 & 0.0           & \textbf{60.0} & \textbf{25.2}
& 64.6 & \textbf{0.0}  & \textbf{70.5} & \textbf{16.7} \\
GPT-5.4-mini$^\ddagger$
& 60.4 & 0.0           & 49.1 & 28.9
& 44.7 & 0.0           & 71.7 & 14.6
& 50.4 & 1.9           & 83.0 & 7.5 \\
Claude Opus 4.7
& 77.1 & 0.0           & 89.0 & 7.8
& 54.8 & 0.0           & 100.0 & 0.0
& 63.3 & \textbf{0.0}  & 100.0 & 0.0 \\
Claude Sonnet 4.6
& \textbf{78.8} & 0.0          & 85.2 & 11.5
& 56.0 & 1.3           & 93.4 & 4.3
& 61.3 & 17.7          & 89.2 & 5.0 \\
Gemini 3.1 Pro
& 75.5 & 0.0           & 73.9 & 18.8
& 56.5 & 0.0           & 85.7 & 8.9
& 58.7 & \textbf{0.0}  & 82.4 & 9.0 \\
Gemini 3.1 Flash-Lite
& 75.4 & 0.0           & 79.1 & 16.8
& \textbf{80.2} & 10.9 & 87.8 & 9.0
& \textbf{70.7} & \textbf{0.0}  & 83.3 & 10.1 \\
Kimi-K2.6
& 68.6 & 0.0           & 90.2 & 6.6
& 56.3 & 0.0           & 84.1 & 9.9
& 63.4 & 2.4           & 92.5 & 4.3 \\
GLM-5.1
& 65.8 & 0.0           & 84.6 & 10.1
& 52.2 & 0.0           & 90.0 & 6.3
& 53.2 & 13.1          & 94.3 & 2.6 \\
Qwen3.6-Plus
& 75.5 & 0.0           & 80.9 & 14.4
& 49.2 & 9.1           & 90.0 & 5.4
& 53.4 & \textbf{0.0}  & 81.2 & 9.6 \\
DeepSeek-V4-Pro
& 57.9 & 0.0           & 68.8 & 17.4
& 43.0 & 10.7          & 86.0 & 7.0
& 55.3 & 7.8           & 83.1 & 8.4 \\
DeepSeek-V4-Flash
& 53.7 & 0.0           & 73.7 & 14.8
& 43.3 & 0.0           & 93.8 & 3.7
& 48.1 & 12.3          & 84.0 & 7.1 \\
MiniMax-M2.7
& 62.0 & 0.0           & 77.1 & 15.7
& 49.0 & 4.4           & 96.9 & 1.8
& 49.4 & 11.7          & 92.5 & 3.1 \\
\bottomrule
\end{tabular}
\caption{\textbf{Multi-paper similarity grouping results} (full content granularity, 85 instances).
Metrics follow the same cascade as Table~\ref{tab:main_pairwise}.
All metrics reported as percentages.
$^\ddagger$GPT-family judged by Gemini-3.1-Pro. \textbf{Bold} marks the best value per column when fewer than half the models tie.}
\label{tab:main_grouping}
\end{table*}

\subsection{Granularity Results}
\label{sec:res:granularity}

\begin{figure}[t]
\centering
\includegraphics[width=0.9\columnwidth]{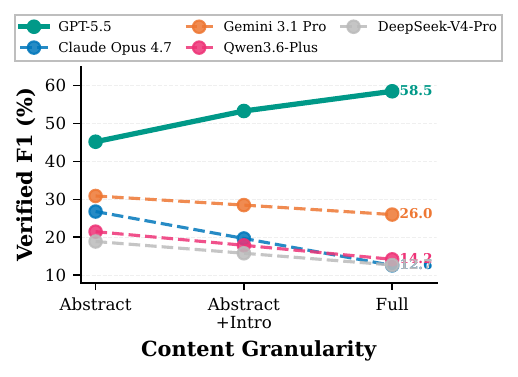}
\caption{Verified F1 across three content granularities for five representative models.}
\label{fig:granularity}
\end{figure}

We rerun pairwise judgment at three content granularities to test whether richer
paper context improves both accuracy and faithfulness. Figure~\ref{fig:granularity}
shows five representative models. Full results appear in Appendix~\ref{app:granularity}.

\paragraph{More context improves raw F1 but reduces Verified F1.}
Raw F1 increases for 13 of 18 models, with a mean gain of 3.9 points. But verified
F1 decreases for 16 of 18 models, with a mean loss of 5.0 points. GPT-5.5 and GPT-5.4-mini are the only models
that maintain or improve Verified F1 with longer content. The pattern suggests
that richer context helps models resolve ambiguous overlap labels, but makes it
harder to keep the rationale tightly grounded.

\paragraph{Longer inputs invite mismatch, not mainly hallucination.}
The degradation comes from unsupported reasoning rather than fabricated spans.
Mismatch Rate rises from 63.6\% on abstracts to 72.6\% on full papers, while
Hallucination Rate remains low at 0.5\% on abstracts and 6.3\% on full papers.
Longer papers give models more plausible evidence to choose from, but much of
that evidence is not logically tied to the stated novelty judgment.

\subsection{Grouping Results}
\label{sec:res:multipaper}

Table~\ref{tab:main_grouping} reports results for the 12 frontier models on 85
multi-paper grouping instances. The task uses the same cascade as pairwise judgment, but each predicted group
must cite evidence from its member papers to justify the shared-dimension judgment.

\paragraph{Grouping amplifies the faithfulness cliff.}
Panel-mean Verified F1 drops to 17\%, 8\%, and 7\% for task, problem, and method.
These scores retain only 38--62\% of the corresponding pairwise values. This
indicates that models lose more faithful reasoning ability when the unit of
judgment shifts from one paper pair to a multi-paper set. GPT-5.5 again leads
all three dimensions, with Verified F1 of 41.7\%, 25.2\%, and 16.7\%.

\paragraph{Shorter evidence spans reduce hallucination but do not solve mismatch.}
Hallucination rates are lower in grouping than in pairwise evaluation, with panel
means of 0.0\%, 3.0\%, and 5.6\%. Grouping tends to produce shorter evidence
spans than pairwise judgment, averaging 34 words compared with 53 words in the
pairwise setting. This leaves less room for fabrication. The main failure is
still mismatch, with panel means of 75\%, 87\%, and 86\%. Appendix~\ref{sec:analysis:rank}
examines the resulting rank drift between pairwise and grouping leaderboards.

\section{Conclusion}
\label{sec:conclusion}

We introduced \textsc{NovGauge}, a benchmark that evaluates LLM novelty judgment with per-dimension labels and cascading faithfulness checks.
Our evaluation of 18 LLMs reveals that accuracy is a misleading metric.
Only 5 of 54 model-dimension pairs retain more than half of their raw F1 after faithfulness filtering.
Multi-paper grouping amplifies this faithfulness cliff, where even GPT-5.5 degrades substantially.
This bottleneck is most severe for method-level novelty, worsens with longer 
paper context, and often arises from a gap between contribution-level reasoning and implementation-level evidence.
Across settings, LLMs also exhibit a conservative similarity bias, tending to reject genuine overlap more often than they invent spurious similarity.
Together, these findings show that reliable novelty assessment requires more faithful links between verdicts, reasons, and evidence.

\section*{Limitations}

\begin{itemize}[leftmargin=*,nosep,itemsep=3pt]
  \item \textbf{Domain scope.} All data derives from ICLR and CS arXiv surveys. Cross-disciplinary generalization is untested.
  
  \item \textbf{LLM judge dependency.} The cascading metric relies on LLM judges (GPT-5.2 and Gemini-3.1-Pro). Researchers without access can substitute another capable model at some cost in comparability.
  
  \item \textbf{Per-dimension negative imbalance.} Negative samples are uneven across dimensions, with the problem dimension having only 33 negatives, yielding wider variance in F1 estimates.
  
  \item \textbf{Reviewer LLM contamination.} A fraction of ICLR 2025--2026 reviews may have been drafted with LLM assistance, though the underlying novelty judgments still reflect human expertise.

  \item \textbf{Retrieval scope.} \textsc{NovGauge} evaluates judgment after candidate papers are retrieved. It can serve as the judgment stage of a retrieval pipeline. Evaluation of the full retrieval process is left to future work.
\end{itemize}

\section*{Ethics Statement}
All data derives from publicly available OpenReview submissions and arXiv papers.
No personally identifiable information beyond author names on public papers is included.
The benchmark is intended for research on evaluation methodology and does not endorse automated replacement of human peer review.

\bibliography{main}

\appendix


\section{Use of AI Assistance}
\label{app:ai_assistance}

Generative AI tools were used to assist with the preparation of this manuscript,
including language polishing, organization of draft material, LaTeX editing, and
consistency checks. AI tools were not used as authors and did not determine the
research questions, benchmark design, experimental results, or scientific
claims. The authors reviewed, edited, and verified all AI-assisted output and
take full responsibility for the content of the paper.

\section{Model and Inference Details}
\label{app:models}

\subsection{Model Coverage}

We evaluate 18 LLMs in three groups. The frontier group contains 12 systems:
GPT-5.5, GPT-5.4-mini, Claude Opus 4.7, Claude Sonnet 4.6, Gemini 3.1 Pro,
Gemini 3.1 Flash-Lite, Kimi-K2.6, GLM-5.1, Qwen3.6-Plus, DeepSeek-V4-Pro,
DeepSeek-V4-Flash, and MiniMax-M2.7. The mid-scale open-weight group contains
Qwen3-8B, Qwen3-32B, Qwen3.5-27B, and Gemma4-31B-IT. The specialised
peer-review group contains CycleReviewer-8B, DeepReviewer-7B,
DeepReviewer-14B, and LLaMA-OpenReviewer-8B.

DeepReviewer-7B and LLaMA-OpenReviewer-8B failed to produce valid structured
outputs and are excluded from quantitative tables. Appendix~\ref{app:failed_models}
reports the failure cases.

\subsection{Inference Settings}

All models are run with their maximum reasoning setting when the API or local
runtime exposes one. Pairwise judgment is evaluated for all 18 models.
Multi-paper grouping is restricted to the 12 frontier models because the task
requires substantially longer context windows.

Papers are rendered at three granularities: abstract only, abstract plus
introduction, and full paper. The corresponding token caps are 1k, 4k, and 16k
per paper. Each input and dimension is decoded three times at temperature 0.6.
Per-family prompt and configuration details appear in Appendix~\ref{app:prompts}.

\section{Metric Definitions}
\label{app:metrics}

This section provides formal definitions for the three cascading metrics introduced 
in Section~\ref{sec:eval_protocol}. All metrics are computed per dimension 
(task, problem, method) and averaged across independent samples.

We first define the shared notation. For a given model and dimension, let:
\begin{itemize}[leftmargin=*,itemsep=2pt]
\item $\text{TP}$ = true positives (pred=similar, label=similar),
\item $\text{FP}$ = false positives (pred=similar, label=dissimilar),
\item $\text{FN}$ = false negatives (pred=dissimilar, label=similar).
\end{itemize}

Among the true positives, the cascading evaluation partitions $\text{TP}$ into 
three mutually exclusive subsets based on Stages 2 and 3:
\begin{itemize}[leftmargin=*,itemsep=2pt]
\item $\text{TP}_h$ = hallucinated (Stage 2 failed: evidence not grounded in paper text),
\item $\text{TP}_m$ = mismatched (Stage 2 passed but Stage 3 failed: reason not supported by evidence),
\item $\text{TP}_v$ = verified (both Stage 2 and Stage 3 passed).
\end{itemize}
By construction, $\text{TP} = \text{TP}_v + \text{TP}_h + \text{TP}_m$.

\subsection{Hallucination Rate}

Hallucination Rate measures the fraction of true-positive predictions whose cited 
evidence is not grounded in the source paper text. An evidence span is considered 
hallucinated if it fails both exact substring match and 4-gram soft match 
(ROUGE-L recall $\geq 0.75$) against the paper content.
\[
\text{Hallucination Rate} = \frac{|\text{TP}_h|}{|\text{TP}|}
\]

\subsection{Mismatch Rate}

Among the non-hallucinated true positives ($\text{TP}_v + \text{TP}_m$), Mismatch
Rate measures the fraction whose stated reason is judged unsupported by the cited
evidence. An external LLM judge makes this determination, using the prompt in
Appendix~\ref{app:prompts}.
\[
\text{Mismatch Rate} = \frac{|\text{TP}_m|}{|\text{TP}_v| + |\text{TP}_m|}
\]

Tables~\ref{tab:main_pairwise} and~\ref{tab:main_grouping} report Hallucination 
Rate and Mismatch Rate computed over true positives only. False-positive predictions 
(pred=similar, label=dissimilar) undergo the same two-stage verification, their 
substantially higher hallucination and mismatch rates are reported separately in 
Figure~\ref{fig:tp_fp_hallucination}.

\subsection{Verified F1}

Verified F1 measures end-to-end performance: a true positive counts toward the 
numerator only if it passes both Stage 2 and Stage 3.
\begin{align}
\text{Verified Precision} &= \frac{\text{TP}_v}{\text{TP}_v + \text{TP}_h + \text{TP}_m + \text{FP}} \\
\text{Verified Recall} &= \frac{\text{TP}_v}{\text{TP}_v + \text{TP}_h + \text{TP}_m + \text{FN}}
\end{align}
Verified F1 is the harmonic mean:
\[
\text{Verified F1} =
\frac{2 \cdot \text{VP} \cdot \text{VR}}{\text{VP} + \text{VR}}
\]
where VP and VR denote Verified Precision and Verified Recall.

Note that the denominator of Verified Precision equals $\text{TP} + \text{FP}$, 
identical to standard Precision. The only difference is that the numerator counts 
only the verified subset $\text{TP}_v$ rather than all $\text{TP}$.

\textbf{Granularity for multi-paper grouping.}
For the grouping evaluation, hallucination and mismatch checks operate at the 
\emph{subgroup} level. Each predicted subgroup is classified as verified, 
hallucinated, or mismatched based on its collective evidence and reasoning. 
However, F1 computation operates at the \emph{pair} level to remain comparable 
with pairwise evaluation. The procedure is as follows.
\begin{enumerate}
  \item Each predicted subgroup $G$ is classified as verified, hallucinated, or
        mismatched.
  \item The subgroup is then converted to its set of within-group paper pairs: 
        $\text{pairs}(G) = \{(i,j) : i, j \in G, i < j\}$.
  \item All pairs from verified subgroups contribute to $\text{TP}_v$ when they
        match ground-truth pairs and to $\text{FP}$ otherwise. Similarly, pairs
        from hallucinated or mismatched subgroups contribute to $\text{TP}_h$ or
        $\text{TP}_m$ when correct and to $\text{FP}$ otherwise.
\end{enumerate}
This design ensures that a single hallucinated evidence span invalidates all pairs 
within that subgroup, reflecting the higher reliability standard for multi-document 
reasoning, while maintaining pair-level F1 comparability with the pairwise evaluation.

\section{ICLR Construction Pipeline Details}
\label{app:iclr_construction}

This appendix provides the complete construction pipeline for the ICLR
review-grounded subset, including extraction rules, verification procedures,
tiering criteria, and quality controls.

\subsection{Candidate Extraction Rules}

\textbf{Novelty-overlap claim definition.}
A candidate novelty-overlap claim is a statement in an official review or author
rebuttal that satisfies both conditions below.
\begin{itemize}
  \item A reviewer asserts that the submission's task, problem, or method is
        already covered by a specific prior work
  \item The authors explicitly concede such a claim in their rebuttal.
\end{itemize}

The extraction step uses GPT-5.5. We prompt it to operate at high recall and capture any
statement that could plausibly indicate overlap, even if the reviewer's phrasing is indirect
or the concern is later refuted by the authors.

\textbf{Extraction unit.}
Each candidate is represented as a pair consisting of a submission and a cited prior work.
The fields are listed below.
\begin{itemize}
  \item The \texttt{submission} field contains the ICLR forum ID and year.
  \item The \texttt{cited\_prior\_work} field contains the paper title, authors, and
        venue as mentioned in the review or rebuttal.
  \item The \texttt{reviewer\_quote} field contains the verbatim text from the review
        that contains the overlap claim.
  \item The \texttt{author\_rebuttal\_quote} field is optional. It contains the
        verbatim text from the rebuttal where the authors acknowledge the concern.
\end{itemize}

Raw extraction yields tens of thousands of noisy candidates per year, including
false positives where reviewers mention prior work for comparison without claiming
overlap.
GPT-5.5 also proposes preliminary dimension labels, and a coding agent using GPT-5.5
separately collects paper metadata. The verification stage revisits the labels.

\subsection{LLM Verification and Dimension Correction}

\textbf{Verification criteria.}
Claude Opus 4.6 then verifies each candidate using the following criteria.
\begin{enumerate}
  \item The claim is genuinely a novelty concern, not merely a citation for
        background or comparison.
  \item The overlap is specific to the cited prior work, not a general statement
        about the field.
  \item The labels for task, problem, and method are correctly assigned
        based on the reviewer's statement.
\end{enumerate}

\textbf{Dimension correction.}
The verification LLM may correct the initial dimension labels if the reviewer's
statement indicates a different type of overlap than initially extracted.
For example, a reviewer statement ``This method is similar to [Prior Work]'' is
labeled as \texttt{method}, while ``This paper addresses the same problem as
[Prior Work]'' is labeled as \texttt{problem}.

\subsection{Prior-Work Identity Resolution}

\textbf{Multi-source resolution chain.}
Each candidate's cited prior work is resolved to a stable identity through a
cascading lookup across the following sources.
\begin{enumerate}
  \item \textbf{arXiv.} Title and author matching against arXiv metadata.
  \item \textbf{Semantic Scholar.} API lookup by title, with fuzzy matching for
        minor variations.
  \item \textbf{OpenReview.} Direct forum ID if the prior work was also submitted
        to ICLR.
  \item \textbf{Publisher pages.} DOI resolution for conference or journal papers.
  \item \textbf{DBLP.} Bibliographic lookup for venue-published papers.
  \item \textbf{Manual web search.} Human fallback for ambiguous cases.
\end{enumerate}

\textbf{Resolution confidence.}
Each resolved prior work receives one of the following confidence levels.
\begin{itemize}
  \item \textbf{High.} Exact match on title and authors, or a confirmed DOI or arXiv ID.
  \item \textbf{Medium.} Fuzzy title match with author overlap, or confirmation from
        a single source.
  \item \textbf{Low.} Ambiguous match or multiple plausible candidates.
\end{itemize}

The final test set retains only pairs with \texttt{high} resolution confidence.

\subsection{Author Concession Verification}

\textbf{Structured signal.}
Author concession is recorded only when the author \emph{explicitly admits the
specific concern} raised by the reviewer.
Generic statements like ``We thank the reviewer for the suggestion'' or ``We will
cite this work'' do not count as concession.

\textbf{Verification protocol.}
The human adjudicator verifies the following conditions.
\begin{enumerate}
  \item The rebuttal quote directly responds to the reviewer's overlap claim.
  \item The rebuttal acknowledges the overlap. For example, it may state ``We agree
        that our method shares similarities with [Prior Work].''
  \item The rebuttal refers to the same prior work the reviewer named.
\end{enumerate}

\subsection{Tier Assignment Rules}

\textbf{Tier-1 criteria.}
A pair is assigned \textbf{Tier-1} only when both conditions hold.
\begin{itemize}
  \item Multiple reviewers independently flag the same prior work. At least two
        distinct reviewers must mention the same paper.
  \item The authors explicitly concede the concern in their rebuttal.
\end{itemize}

\textbf{Tier-2 criteria.}
A pair is assigned \textbf{Tier-2} when either condition holds.
\begin{itemize}
  \item Only one reviewer flags the prior work.
  \item Multiple reviewers flag it but the authors do not concede.
\end{itemize}

Pairs with neither signal are dropped from the benchmark draft. A pair has neither
signal when only one reviewer flags the prior work and the authors do not concede.

\subsection{Human Adjudication Protocol}

\textbf{Partition and audit procedure.}
Three PhD-level annotators adjudicate every candidate pair. All three have research
experience in NLP or ML. They follow the protocol below.
\begin{enumerate}
  \item \textbf{Partition.} Each pair is assigned to one primary annotator.
  \item \textbf{Label.} The primary annotator labels the pair, verifying prior-work
        identity, dimension labels, and tier assignment.
  \item \textbf{Audit.} The other two annotators independently audit the primary
        annotator's decision.
  \item \textbf{Resolve.} Disagreements are resolved through discussion among all
        three annotators.
\end{enumerate}

\textbf{Adjudication checklist.}
For each pair, the primary annotator verifies the following conditions.
\begin{itemize}
  \item The cited prior work is correctly resolved to a stable identity.
  \item The reviewer quotes anchor to that specific paper.
  \item The labels for task, problem, and method are supported by the raw
        forum evidence.
  \item The tier assignment follows the Tier-1 and Tier-2 rules and reflects the
        number of reviewers and any author concession.
\end{itemize}

\textbf{Adjudication outcomes.}
Adjudication has one of the following outcomes.
\begin{itemize}
  \item \textbf{Accept.} The pair is retained with the verified labels.
  \item \textbf{Correct.} The pair is retained, but its dimension labels or tier are
        corrected.
  \item \textbf{Reject.} The pair is excluded when the prior work is misidentified or
        the claim is not genuinely about novelty.
\end{itemize}

\textbf{Consistency audit by dimension.}
For the consistency audit in Section~\ref{sec:data_construction}, we sampled 30 items
with 10 from each dimension. Three annotators independently reviewed each item,
giving 90 reviews in total. Table~\ref{tab:dim_agreement} reports 93.3\%
overall agreement. Disagreements mainly involved the boundary between task and
problem and differences in method granularity. All were resolved through discussion.

\begin{table}[t]
\centering\small
\begin{tabular}{lrrr}
\toprule
\textbf{Dimension} & \textbf{Reviews} & \textbf{Disagree} & \textbf{Agreement} \\
\midrule
Task    & 30 & 1 & 96.7\% \\
Problem & 30 & 3 & 90.0\% \\
Method  & 30 & 2 & 93.3\% \\
\midrule
Overall & 90 & 6 & 93.3\% \\
\bottomrule
\end{tabular}
\caption{Label consistency audit by dimension. Three annotators independently
reviewed 30 sampled items, with 10 from each dimension, giving 90 reviews in total.
Agreement is computed against the original annotations.}
\label{tab:dim_agreement}
\end{table}

\subsection{Final Selection Criteria}

The released test set retains pairs that meet all of the following conditions.
\begin{itemize}
  \item \texttt{tier == "Tier-1"}
  \item \texttt{human\_adjudicated == true}
  \item \texttt{adjudication\_source == "human"}. Agent-adjudicated entries are excluded.
  \item \texttt{overlap.is\_valid == true}
  \item \texttt{prior\_resolved == true}
  \item \texttt{prior\_resolution\_confidence == "high"}
  \item \texttt{reviewer\_evidence.quotes} is non-empty
  \item Both submission and prior work have parsed Markdown content
\end{itemize}

This strict filtering ensures that every pair in the test set is grounded in
convergent human evidence from multiple reviewers and an author concession,
and is anchored to a correctly identified prior work.

\subsection{Year Distribution and Temporal Skew}

\begin{table}[t]
\centering\small
\begin{tabular}{lrrrr}
\toprule
\textbf{Dimension} & \textbf{2023} & \textbf{2024} & \textbf{2025} & \textbf{2026} \\
\midrule
method  & 11 & 7  & 10 & 169 \\
problem & 17 & 13 & 11 & 60  \\
task    & 1  & 0  & 0  & 3   \\
\midrule
\textbf{Total pairs} & 19 & 13 & 15 & 189 \\
\bottomrule
\end{tabular}
\caption{ICLR dimension labels by year. Pairs with multi-dimension labels appear
in more than one row. The total pair count is the unique pair count per year.}
\label{tab:iclr_dim_by_year}
\end{table}

The pairs concentrate in ICLR 2026. It contributes 189 of 236 pairs, or 80\%.
This concentration reflects two \emph{construction} factors rather than a change
in reviewer behavior.
\begin{enumerate}
  \item ICLR 2026 contributes by far the most raw OpenReview forums to the
        extraction pipeline.
  \item The benchmark's human-review effort prioritized a test-heavy 2026 split,
        while 2025 adjudication was directed mainly toward a training split that
        is not released with this benchmark.
\end{enumerate}

The year distribution should therefore not be read as an estimate of the true
per-year novelty-overlap rate in ICLR submissions.

\textbf{Evaluation by year.}
Because ICLR contributes positive pairs only, F1 and Verified F1 are not comparable
across ICLR years. We therefore report metrics based on positive pairs only.
The method dimension has 169 pairs from 2026 and 28 pairs from earlier years.
The problem dimension has 60 pairs from 2026 and 41 pairs from earlier years.
We omit the task dimension because it has only 3 pairs from 2026 and 1 pair from
earlier years. Table~\ref{tab:year_metrics} averages the metrics over input settings
and evaluated models. The values are similar across years.

\begin{table}[t]
\centering\small
\setlength{\tabcolsep}{1.5pt}
\begin{tabular}{lrrrr}
\toprule
\textbf{Dimension, year} & \shortstack{\textbf{Recall}\\\textbf{(\%)}} & \shortstack{\textbf{V-Recall}\\\textbf{(\%)}} & \shortstack{\textbf{Mismatch}\\\textbf{(\%)}} & \shortstack{\textbf{Halluc.}\\\textbf{(\%)}} \\
\midrule
Method, 2026          & 38.8 & 12.6 & 68.4 & 4.8 \\
Method, earlier years & 36.7 & 9.4  & 76.2 & 4.6 \\
Problem, 2026          & 49.2 & 16.3 & 67.4 & 1.4 \\
Problem, earlier years & 49.1 & 17.9 & 64.6 & 1.7 \\
\bottomrule
\end{tabular}
\caption{ICLR metrics for positive pairs, stratified by year. Each row reports one
dimension and year group. Values are percentages averaged over input settings and
models. The task dimension is omitted because it has only 3 pairs from 2026 and 1
pair from earlier years.}
\label{tab:year_metrics}
\end{table}

We also compared the full ICLR model ranking with the ranking on data from earlier
years. Table~\ref{tab:year_ranking} shows Spearman correlations of 0.87--0.98
across input settings and metrics. These values indicate stable rankings after
removing 2026.

\begin{table}[t]
\centering\small
\setlength{\tabcolsep}{5pt}
\begin{tabular}{lccc}
\toprule
\textbf{Metric} & \textbf{Abstract} & \textbf{Abstract and intro} & \textbf{Full} \\
\midrule
Recall          & 0.97 & 0.93 & 0.90 \\
Verified-Recall & 0.96 & 0.87 & 0.97 \\
Mismatch        & 0.95 & 0.90 & 0.98 \\
\bottomrule
\end{tabular}
\caption{Spearman correlation between the full ICLR ranking and the pre-2026 ranking.
Values are reported for each input setting and metric.}
\label{tab:year_ranking}
\end{table}

\subsection{Construction Funnel Summary}

\begin{table}[t]
\centering\small
\setlength{\tabcolsep}{3pt}
\begin{tabular}{lr}
\toprule
\textbf{Stage} & \textbf{Count} \\
\midrule
Raw OpenReview forums, 2023 to 2026 & 42{,}682 \\
Candidate pairs from extraction     & 66{,}332 \\
Pairs after LLM verification        & $\sim$10{,}000 \\
Tier 1 pairs after filtering        & $\sim$2{,}000 \\
Pairs after human adjudication      & 236 \\
\midrule
\textbf{Reduction ratio}                 & about 280 to 1 \\
\bottomrule
\end{tabular}
\caption{ICLR construction funnel from raw forums to final test set.}
\label{tab:iclr_funnel}
\end{table}

The 236 released pairs are a deliberately selected, content-available benchmark
subset, not an exhaustive set of all novelty-overlap concerns in ICLR.
This funnel reflects high-precision filtering for evaluation rather than
prevalence estimation.

\section{Survey Construction Pipeline Details}
\label{app:survey_construction}

This appendix provides the complete construction pipeline for the survey
co-citation subset, including survey filtering rules, sentence extraction patterns,
annotation protocols, and cross-group sampling constraints.

\subsection{Survey Collection Rules}

\textbf{Precision-oriented filter.}
A paper qualifies as a candidate survey when one of the following conditions holds.
\begin{itemize}
  \item Its title contains \textit{survey}, \textit{review}, or \textit{tutorial}.
  \item Its abstract opens with an explicit survey or review declaration. For example,
        it may begin ``This survey reviews...'' or ``We present a comprehensive review of...''.
\end{itemize}

\textbf{False positive exclusion.}
The filter excludes papers whose title contains any of the following phrases.
\begin{itemize}
  \item \textit{peer review}, as in ``Peer Review Analysis''
  \item \textit{code review}, as in ``Automated Code Review''
  \item \textit{literature review} in a non-survey context, as in ``A New Method
        with Literature Review''
\end{itemize}

This precision-first design prioritizes high-quality survey papers over exhaustive
coverage.

\subsection{Sentence Extraction Rules}

\textbf{Co-citation requirement.}
A sentence qualifies as a candidate if it cites two or more papers.
Citations are identified in standard forms such as \texttt{[Author et al., Year]},
\texttt{(Author et al., Year)}, \texttt{\textbackslash cite\{...\}}, and numbered
references such as \texttt{[1, 2, 3]}.

\textbf{Similarity cues.}
The sentence must contain at least one of the following explicit similarity cues.
\begin{itemize}
  \item \textit{similar}, or \textit{share the same task, problem, method, or approach}
  \item \textit{address the same problem}, \textit{follow the same approach}
  \item \textit{both papers tackle}, \textit{fall into the same}, or \textit{belong to
        the same}
  \item \textit{group of methods}, \textit{family of approaches}
\end{itemize}

\textbf{Anti-patterns.}
The extractor excludes sentences containing the following technical uses of ``similarity''.
\begin{itemize}
  \item \textit{cosine similarity}, \textit{semantic similarity score},
        \textit{similarity metric}
  \item \textit{similarity function}, \textit{similarity measure},
        \textit{similarity threshold}
  \item \textit{Jaccard similarity}, \textit{edit distance similarity}
\end{itemize}

This ensures that only sentences asserting paper-level similarity are kept, not
sentences discussing similarity as a technical concept.

\subsection{LLM Pre-screening Protocol}

The survey pre-screening model is Gemini 3.1 Pro, the only LLM in this pipeline. It
labels sentences selected by the rule-based extractor, and human annotators confirm or
revise each grouping.

\textbf{Input format.}
For each candidate sentence, the LLM receives the following information.
\begin{itemize}
  \item The co-citation sentence (verbatim)
  \item The list of cited papers (titles and authors)
  \item The surrounding context (previous and next sentences)
\end{itemize}

\textbf{Output format.}
The LLM proposes a three-dimensional similarity grouping in the following JSON format.
\begin{verbatim}
{
  "task": [[p1, p2], [p3]],
  "problem": [[p1, p3]],
  "method": [[p2, p3]]
}
\end{verbatim}

Each dimension contains zero or more similarity groups.
Papers not grouped with any other paper are omitted.

\subsection{Human Verification Protocol}

\textbf{Two-outcome protocol.}
For each candidate sentence, the assigned annotator performs the following steps.
\begin{enumerate}
  \item Reads the co-citation sentence, the grouping proposed by the LLM, and the cited
          papers' full text alongside one another.
  \item Decides whether to \textbf{accept} the LLM labels or \textbf{revise} them.
\end{enumerate}

\textbf{Explicit revision.}
If any dimension label is judged incorrect or incomplete, the annotator writes an
explicit correction in the \texttt{human\_revise} field. The correction specifies
the following.
\begin{itemize}
  \item Which papers are similar on which dimension
  \item Which papers are \emph{not} similar when the LLM incorrectly grouped them
\end{itemize}

\textbf{Implicit approval.}
If no correction is needed, the annotator records a non-empty
\texttt{human\_judgment} entry such as ``Verified''. This signals that all three LLM
labels are accepted.

\textbf{Audit.}
The other two annotators independently audit each decision.
Disagreements are resolved through discussion.

\textbf{Final label merge rule.}
The labels for each verified sentence are merged as follows.
\begin{itemize}
  \item If \texttt{human\_revise} is non-empty, its per-dimension labels replace
        the LLM output in full.
  \item Otherwise, the \texttt{llm\_judgment} labels are adopted as-is.
\end{itemize}

\subsection{Cross-Group Sampling Constraints}

\textbf{Dimension-specific sampling.}
For each dimension $d \in \{\text{task}, \text{problem}, \text{method}\}$, we
sample negative pairs only from sentences where the annotators verified at least
one similarity group in dimension $d$.

\textbf{Sampling procedure.}
For each verified sentence with $k \geq 2$ similarity groups in dimension $d$, we do
the following.
\begin{enumerate}
  \item Enumerate all pairs $(p_i, p_j)$ where $p_i$ is in group $g_1$ and $p_j$
        is in group $g_2 \neq g_1$.
  \item Label each such pair as \texttt{not\_similar} in dimension $d$.
\end{enumerate}

\textbf{Conflict handling.}
If a cross-group pair $(p_i, p_j)$ appears as a positive in dimension $d$ elsewhere
in the survey corpus, we exclude it from the negative set. This occurs when the same
two papers are grouped together in another sentence.

\textbf{Negatives in multiple dimensions.}
A single pair may carry negative labels in multiple dimensions if the survey author
placed the two papers into distinct groups across multiple dimensions.
56 of the 156 negative pairs carry labels in more than one dimension.

\subsection{Per-Survey Contribution Distribution}

\begin{table}[t]
\centering\small
\begin{tabular}{lrr}
\toprule
\textbf{Survey rank} & \textbf{Sentences} & \textbf{Cumulative \%} \\
\midrule
Top-1  & 12 & 10.2\% \\
Top-2  & 7  & 16.1\% \\
Top-3  & 7  & 22.0\% \\
Top-5  & 14 & 33.9\% \\
Top-10 & 24 & 54.2\% \\
\midrule
Remaining 27 surveys & 54 & 100.0\% \\
\bottomrule
\end{tabular}
\caption{Per-survey contribution to the 118 verified sentences. The top-10
contributing surveys account for 54\% of all sentences.}
\label{tab:survey_concentration}
\end{table}

The per-survey skew reflects both the precision-first design of the extractor,
which favors surveys with explicit similarity statements, and the natural variation
in how surveys structure their related-work prose.
We treat the resulting concentration as a known limitation.

\section{Evidence Verification Technical Details}
\label{app:evidence_verification}

This appendix provides the complete technical specification for the hallucination
check in Stage 2 of the cascading evaluation protocol.

\subsection{Normalization Rules}

Before exact substring matching, both the cited evidence span and the source paper
text undergo the following normalizations:

\textbf{Ligature unification.}
Unicode ligatures are expanded to their component characters:
\begin{itemize}
  \item \texttt{ﬁ} (U+FB01) $\rightarrow$ \texttt{fi}
  \item \texttt{ﬂ} (U+FB02) $\rightarrow$ \texttt{fl}
  \item \texttt{ﬀ} (U+FB00) $\rightarrow$ \texttt{ff}
  \item \texttt{ﬃ} (U+FB03) $\rightarrow$ \texttt{ffi}
  \item \texttt{ﬄ} (U+FB04) $\rightarrow$ \texttt{ffl}
\end{itemize}

\textbf{Typographic quote and dash normalization.}
Typographic punctuation is normalized to ASCII equivalents:
\begin{itemize}
  \item \texttt{"} (U+201C), \texttt{"} (U+201D) $\rightarrow$ \texttt{"}
  \item \texttt{'} (U+2018), \texttt{'} (U+2019) $\rightarrow$ \texttt{'}
  \item \texttt{–} (U+2013, en-dash), \texttt{—} (U+2014, em-dash) $\rightarrow$
        \texttt{-}
\end{itemize}

\textbf{Soft-hyphen removal.}
Soft hyphens (U+00AD) are removed entirely.

\textbf{PDF line-wrap hyphenation handling.}
Hyphens at line breaks are conditionally removed:
\begin{itemize}
  \item If a word ends with \texttt{-\textbackslash n} and the next line starts
        with a lowercase letter, the hyphen and newline are removed.
  \item Otherwise, the hyphen is retained.
\end{itemize}

\textbf{Whitespace collapse.}
All sequences of whitespace (spaces, tabs, newlines) are collapsed to a single space.

\subsection{Character-Level 4-Gram Recall}

For spans that fail exact substring matching, we compute character-level 4-gram
recall as follows:

\textbf{4-gram extraction.}
For a string $s$ of length $n$, the set of character 4-grams is:
\[
\text{4grams}(s) = \{ s[i:i{+}4] \mid 0 \leq i \leq n{-}4 \}
\]

\textbf{Recall computation.}
For a cited evidence span $e$ and source paper text $p$:
\[
\text{4gram-recall}(e, p) = \frac{|\text{4grams}(e) \cap \text{4grams}(p)|}{|\text{4grams}(e)|}
\]

\textbf{Soft match threshold.}
A sentence is admitted as a soft match if its 4-gram recall is at least 0.75.

\textbf{Rationale.}
This soft match catches PDF parsing artifacts (e.g., \texttt{``per forms''}
matching \texttt{``performs''}) and benign rewordings (e.g., punctuation changes,
minor typos) without admitting unrelated content.
The 0.75 threshold was chosen empirically to balance precision and recall on a
held-out validation set of 50 manually labeled evidence spans.

\begin{table*}[t]
\centering\small
\setlength{\tabcolsep}{4pt}
\begin{tabular}{lcccccc}
\toprule
& \multicolumn{2}{c}{\textbf{Task}}
& \multicolumn{2}{c}{\textbf{Problem}}
& \multicolumn{2}{c}{\textbf{Method}} \\
\cmidrule(lr){2-3} \cmidrule(lr){4-5} \cmidrule(lr){6-7}
\textbf{Model} & F1 & VF1 & F1 & VF1 & F1 & VF1 \\
\midrule
\multicolumn{7}{l}{\textsc{Frontier}} \\
GPT-5.5$^\ddagger$
  & $84.8_{\pm 0.5}$ & $71.5_{\pm 1.6}$ & $69.4_{\pm 0.7}$ & $61.0_{\pm 3.1}$ & $78.7_{\pm 1.0}$ & $43.1_{\pm 2.8}$ \\
GPT-5.4-mini$^\ddagger$
  & $68.1_{\pm 0.4}$ & $52.1_{\pm 2.1}$ & $56.7_{\pm 1.8}$ & $37.3_{\pm 3.0}$ & $60.8_{\pm 1.0}$ & $29.2_{\pm 1.8}$ \\
Claude Opus 4.7
  & $77.7_{\pm 0.4}$ & $18.2_{\pm 1.5}$ & $66.2_{\pm 0.9}$ & $10.5_{\pm 1.0}$ & $72.2_{\pm 0.3}$ & $9.1_{\pm 1.7}$ \\
Claude Sonnet 4.6
  & $84.3_{\pm 0.8}$ & $18.4_{\pm 1.5}$ & $66.1_{\pm 0.7}$ & $13.4_{\pm 1.8}$ & $70.4_{\pm 0.6}$ & $8.6_{\pm 2.5}$ \\
Gemini 3.1 Pro
  & $84.5_{\pm 0.5}$ & $33.5_{\pm 1.4}$ & $77.1_{\pm 1.3}$ & $24.7_{\pm 4.4}$ & $78.0_{\pm 0.8}$ & $19.9_{\pm 1.8}$ \\
Gemini 3.1 Flash-Lite
  & $85.8_{\pm 0.4}$ & $22.8_{\pm 2.6}$ & $89.2_{\pm 0.6}$ & $22.9_{\pm 2.1}$ & $81.9_{\pm 0.2}$ & $15.5_{\pm 2.7}$ \\
Kimi-K2.6
  & $87.6_{\pm 0.4}$ & $19.9_{\pm 1.6}$ & $76.5_{\pm 1.8}$ & $25.7_{\pm 2.0}$ & $64.6_{\pm 1.5}$ & $12.9_{\pm 3.2}$ \\
GLM-5.1
  & $71.2_{\pm 2.2}$ & $24.0_{\pm 2.3}$ & $69.3_{\pm 1.7}$ & $25.3_{\pm 2.5}$ & $57.4_{\pm 0.8}$ & $9.4_{\pm 0.3}$ \\
Qwen3.6-Plus
  & $83.0_{\pm 0.4}$ & $22.0_{\pm 1.3}$ & $74.9_{\pm 1.1}$ & $12.3_{\pm 0.8}$ & $63.0_{\pm 1.3}$ & $8.2_{\pm 0.7}$ \\
DeepSeek-V4-Pro
  & $61.9_{\pm 0.2}$ & $20.0_{\pm 2.4}$ & $60.3_{\pm 1.4}$ & $11.4_{\pm 1.2}$ & $51.2_{\pm 0.8}$ & $6.7_{\pm 1.5}$ \\
DeepSeek-V4-Flash
  & $66.3_{\pm 0.6}$ & $18.1_{\pm 0.7}$ & $60.0_{\pm 1.3}$ & $3.7_{\pm 1.3}$ & $61.3_{\pm 1.7}$ & $5.4_{\pm 1.6}$ \\
MiniMax-M2.7
  & $73.2_{\pm 0.9}$ & $9.9_{\pm 2.1}$ & $73.3_{\pm 1.4}$ & $4.4_{\pm 1.2}$ & $64.3_{\pm 0.8}$ & $5.5_{\pm 2.2}$ \\
\midrule
\multicolumn{7}{l}{\textsc{Mid-scale open-weight}} \\
Qwen3-8B
  & $70.3_{\pm 3.0}$ & $9.6_{\pm 2.4}$ & $53.4_{\pm 5.1}$ & $3.9_{\pm 0.5}$ & $48.1_{\pm 2.8}$ & $4.3_{\pm 1.4}$ \\
Qwen3-32B
  & $61.0_{\pm 1.7}$ & $14.5_{\pm 2.5}$ & $56.6_{\pm 2.2}$ & $16.2_{\pm 3.2}$ & $49.3_{\pm 0.8}$ & $8.9_{\pm 2.0}$ \\
Qwen3.5-27B
  & $76.2_{\pm 1.3}$ & $14.3_{\pm 0.7}$ & $76.7_{\pm 2.0}$ & $10.6_{\pm 2.0}$ & $67.1_{\pm 0.9}$ & $8.2_{\pm 2.7}$ \\
Gemma4-31B-IT
  & $73.7_{\pm 0.3}$ & $34.9_{\pm 1.2}$ & $66.2_{\pm 0.8}$ & $31.7_{\pm 1.4}$ & $64.1_{\pm 0.8}$ & $19.2_{\pm 1.6}$ \\
\midrule
\multicolumn{7}{l}{\textsc{Specialised}} \\
CycleReviewer-8B
  & $20.6_{\pm 1.9}$ & $2.4_{\pm 0.0}$ & $22.5_{\pm 3.8}$ & $2.9_{\pm 2.2}$ & $10.0_{\pm 1.1}$ & $0.5_{\pm 0.4}$ \\
DeepReviewer-14B
  & $60.7_{\pm 2.9}$ & $15.9_{\pm 0.9}$ & $73.1_{\pm 2.2}$ & $15.7_{\pm 1.5}$ & $41.9_{\pm 1.5}$ & $9.7_{\pm 1.3}$ \\
\bottomrule
\end{tabular}
\caption{Per-model raw F1 and Verified F1 with cross-sample standard
deviations for pairwise judgment, full content granularity.
Values are mean $\pm$ std across the $n{=}3$ inference samples (same
samples used to compute Table~\ref{tab:main_pairwise}).
$^\ddagger$GPT-family judged by Gemini-3.1-Pro.}
\label{tab:detail_pairwise_std}
\end{table*}

\section{Dataset Schema and Samples}
\label{app:schema}

The benchmark comprises three dataset types corresponding to the two evaluation tasks described in \S\ref{sec:task}.

\subsection{Pairwise Positive Pairs}

Confirmed positive pairs from ICLR submissions citing prior work. Each pair includes the submitted paper, the cited prior work, and human-annotated overlap dimensions.

\textbf{Key fields:}
\begin{itemize}[leftmargin=*,itemsep=1pt]
\item \texttt{submitted\_paper}, \texttt{prior\_work}: paper metadata (title, abstract, etc.)
\item \texttt{similar\_dimensions}: \texttt{["task"]}, \texttt{["problem"]}, \texttt{["method"]}, or combinations
\item \texttt{overlap}: human-written overlap summary
\end{itemize}

\subsection{Cross-Group Negative Pairs}

Negative pairs from papers in different groups within survey sentences. Used to construct hard negatives where papers share context but differ on specific dimensions.

\textbf{Key fields:}
\begin{itemize}[leftmargin=*,itemsep=1pt]
\item \texttt{paper\_a}, \texttt{paper\_b}: paper metadata
\item \texttt{negative\_dimension}: the dimension on which they differ (\texttt{"task"}, \texttt{"problem"}, or \texttt{"method"})
\item \texttt{labels}: per-dimension binary labels (0 = dissimilar)
\end{itemize}

\begin{table*}[!t]
\centering\small
\setlength{\tabcolsep}{2.5pt}
\begin{tabular}{l ccc ccc ccc >{\columncolor{blue!8}}c>{\columncolor{blue!8}}c>{\columncolor{blue!8}}c}
\toprule
& \multicolumn{3}{c}{\textbf{F1}$\uparrow$}
& \multicolumn{3}{c}{\textbf{Hal}$\downarrow$}
& \multicolumn{3}{c}{\textbf{Mis}$\downarrow$}
& \multicolumn{3}{c}{\textbf{VF1}$\uparrow$} \\
\cmidrule(lr){2-4} \cmidrule(lr){5-7} \cmidrule(lr){8-10} \cmidrule(lr){11-13}
\textbf{Model} & abs & a+i & full & abs & a+i & full & abs & a+i & full & abs & a+i & full \\
\midrule
\rowcolor{gray!10}
\multicolumn{13}{l}{\textsc{Frontier}} \\
GPT-5.5$^\ddagger$
& 71.0 & 75.7 & 77.6 & \textbf{0.1} & \textbf{0.0} & \textbf{0.0} & \textbf{38.3} & \textbf{30.3} & \textbf{24.4} & \textbf{45.2} & \textbf{53.3} & \textbf{58.5} \\
GPT-5.4-mini$^\ddagger$
& 58.9 & 58.5 & 61.8 & 0.3 & 0.0 & 0.4 & 44.1 & 35.3 & 36.2 & 34.4 & 38.8 & 39.5 \\
Claude Opus 4.7
& 73.0 & 70.4 & 72.0 & 0.0 & 0.0 & 1.9 & 63.8 & 72.9 & 82.5 & 26.8 & 19.2 & 12.6 \\
Claude Sonnet 4.6
& 74.8 & 74.0 & 73.6 & 0.0 & 0.7 & 8.4 & 72.2 & 73.6 & 80.5 & 21.6 & 19.6 & 13.5 \\
Gemini 3.1 Pro
& 71.7 & 75.6 & 79.9 & 0.0 & 0.0 & 1.8 & 57.7 & 58.2 & 67.1 & 30.9 & 31.9 & 26.0 \\
Gemini 3.1 Flash-Lite
& 65.3 & \textbf{79.6} & \textbf{85.7} & 0.0 & 1.9 & 5.3 & 62.8 & 69.9 & 75.0 & 25.2 & 23.6 & 20.4 \\
Kimi-K2.6
& \textbf{79.9} & 54.6 & 76.3 & 0.1 & 0.0 & 2.9 & 75.4 & 46.3 & 73.9 & 19.9 & 30.1 & 19.5 \\
GLM-5.1
& 63.5 & 64.7 & 65.9 & 0.0 & 0.0 & 9.6 & 63.3 & 63.0 & 68.9 & 23.5 & 24.2 & 19.6 \\
Qwen3.6-Plus
& 69.4 & 72.8 & 73.6 & 0.0 & 0.0 & 2.1 & 69.8 & 71.9 & 81.0 & 21.5 & 20.7 & 14.2 \\
DeepSeek-V4-Pro
& 56.7 & 56.6 & 57.8 & 0.0 & 0.1 & 8.6 & 67.2 & 64.4 & 77.1 & 18.9 & 20.2 & 12.7 \\
DeepSeek-V4-Flash
& 55.4 & 57.1 & 62.6 & 0.1 & 0.0 & 5.8 & 69.3 & 76.1 & 85.2 & 17.6 & 13.8 & 9.1 \\
MiniMax-M2.7
& 58.7 & 65.5 & 70.3 & 0.2 & 0.5 & 5.0 & 75.8 & 85.3 & 90.1 & 14.1 & 9.3 & 6.6 \\
\midrule
\rowcolor{gray!10}
\multicolumn{13}{l}{\textsc{Mid-scale open-weight}} \\
Qwen3-8B
& 51.0 & 52.5 & 57.3 & 0.2 & 0.1 & 6.8 & 70.5 & 77.7 & 89.3 & 13.9 & 11.2 & 5.9 \\
Qwen3-32B
& 46.2 & 51.4 & 55.6 & 0.5 & 0.6 & 5.6 & 61.8 & 70.4 & 75.3 & 17.5 & 15.2 & 13.2 \\
Qwen3.5-27B
& 70.0 & 70.3 & 73.3 & 0.1 & 0.3 & 2.1 & 71.3 & 76.8 & 84.8 & 20.0 & 16.2 & 11.0 \\
Gemma4-31B-IT
& 63.3 & 63.8 & 68.0 & 0.0 & 0.0 & 1.8 & 45.9 & 52.5 & 57.7 & 35.1 & 30.7 & 28.6 \\
\midrule
\rowcolor{gray!10}
\multicolumn{13}{l}{\textsc{Specialised}} \\
CycleReviewer-8B
& 29.4 & 35.3 & 17.7 & 4.8 & 7.2 & 29.7 & 72.2 & 78.8 & 86.4 & 8.4 & 7.7 & 1.9 \\
DeepReviewer-14B
& 59.1 & 63.1 & 58.6 & 2.2 & 2.1 & 16.1 & 62.8 & 63.6 & 71.8 & 21.0 & 22.6 & 13.8 \\
\bottomrule
\end{tabular}
\caption{\textbf{Content granularity ablation.} Metrics averaged over the task,
problem, and method dimensions. \textbf{F1} measures binary prediction accuracy.
\textbf{Hal} denotes Hallucination Rate, \textbf{Mis} denotes Mismatch Rate, and
\textbf{VF1} denotes Verified F1. All metrics are percentages.
The columns are \emph{abs} for abstract only, \emph{a+i} for abstract and intro, and
\emph{full} for the full paper. $^\ddagger$GPT-family models are judged by
Gemini-3.1-Pro. \textbf{Bold} indicates the best value for each content granularity
and metric.}
\label{tab:granularity_ablation}
\end{table*}


\subsection{Multi-paper Grouping Tasks}

Sets of 3--4 papers from survey sentences with ground-truth groupings. Models must partition papers by shared task, problem, or method.

\textbf{Key fields:}
\begin{itemize}[leftmargin=*,itemsep=1pt]
\item \texttt{papers}: list of paper metadata
\item \texttt{ground\_truth}: \texttt{\{task\_groups, problem\_groups, method\_groups\}}, each a list of paper-index lists (e.g., \texttt{[[1,2], [3]]} means papers 1,2 form one group, paper 3 is alone)
\item \texttt{eval\_dims}: dimensions to evaluate on this instance
\end{itemize}

\textbf{Example structure:}
\begin{lstlisting}[basicstyle=\scriptsize\ttfamily]
{
  "papers": [
    {"title": "...", "abstract": "..."},
    {"title": "...", "abstract": "..."},
    {"title": "...", "abstract": "..."}
  ],
  "ground_truth": {
    "task_groups": [[0,1,2]],
    "method_groups": [[0,1], [2]]
  },
  "eval_dims": ["task", "method"]
}
\end{lstlisting}

\subsection{Paper Type and Domain Coverage}
\label{app:coverage}

We classify 679 papers (deduplicated by title)
by contribution type and scientific domain using titles, abstracts, and source
descriptions. Tables~\ref{tab:paper_types} and~\ref{tab:domains} report the
overall and source-specific distributions. Method papers account for 73.0\%, while
the remaining papers include analysis, benchmarks, theory, and surveys. Computer
science accounts for 86.6\% of the papers. The survey source is more diverse than
ICLR. Computer science accounts for 66.5\% of the survey papers and 96.5\% of the
ICLR papers.

Among 619 pairs, 19.5\% span different paper types and 11.0\% span different
scientific domains. The corresponding rates are 19.5\% and 7.2\% for ICLR, and
19.6\% and 13.3\% for the survey source.

\begin{table}[t]
\centering\small
\setlength{\tabcolsep}{5pt}
\begin{tabular}{lrrrr}
\toprule
\textbf{Type} & \textbf{\#} & \textbf{All} & \textbf{ICLR} & \textbf{Survey} \\
\midrule
Method    & 496 & 73.0 & 74.3 & 70.5 \\
Analysis  & 77  & 11.3 & 11.2 & 11.6 \\
Benchmark & 45  & 6.6  & 7.3  & 5.4  \\
Theory    & 38  & 5.6  & 6.8  & 3.1  \\
Survey    & 23  & 3.4  & 0.4  & 9.4  \\
\bottomrule
\end{tabular}
\caption{Contribution-type distribution over 679 papers deduplicated by title. The \textbf{\#} column
gives the count. The other columns are percentages overall and within each source.}
\label{tab:paper_types}
\end{table}

\begin{table}[t]
\centering\small
\setlength{\tabcolsep}{5pt}
\begin{tabular}{lrrrr}
\toprule
\textbf{Domain} & \textbf{\#} & \textbf{All} & \textbf{ICLR} & \textbf{Survey} \\
\midrule
Computer Science     & 588 & 86.6 & 96.5 & 66.5 \\
Medical \& Life Sci. & 38  & 5.6  & 2.2  & 12.5 \\
Elec.\ Eng.\ \& Sys. & 26  & 3.8  & 0.2  & 11.2 \\
Mathematics          & 14  & 2.1  & 0.7  & 4.9  \\
Earth \& Env.        & 9   & 1.3  & 0.4  & 3.1  \\
Social Sciences      & 4   & 0.6  & 0.0  & 1.8  \\
\bottomrule
\end{tabular}
\caption{Domain distribution over 679 papers deduplicated by title. The \textbf{\#} column gives the
count. The other columns are percentages overall and within each source.}
\label{tab:domains}
\end{table}

\section{Detailed Per-Sample Results}
\label{app:detailed_results}

This section provides per-sample variability estimates for the main pairwise
evaluation results.
Table~\ref{tab:detail_pairwise_std} reports raw F1 and Verified F1 with
cross-sample standard deviations for the \textit{full} content granularity, the
default setting in Table~\ref{tab:main_pairwise}.
Each standard deviation is computed across three independent inference runs at
temperature 0.6 on the same test set, reflecting output variability due to
sampling randomness.

Hallucination Rate and Mismatch Rate standard deviations are omitted for space
and are available in the released per-sample raw counts.
The reported standard deviations are typically small, at 0.5--3.0 percentage
points, confirming that the benchmark's conclusions are robust to sample
variation.

\textbf{Dimension coverage and accuracy by source.}
Table~\ref{tab:dim_counts} gives per-dimension label counts, and
Table~\ref{tab:source_controlled} separates accuracy by source. Since ICLR contributes
positives only, we compare both labels within the survey source. Positive accuracy is
44--59\% versus 90--97\% for negatives. ICLR positive accuracy is similar on the
shared dimensions.

\begin{table}[t]
\centering\small
\begin{tabular}{lrr}
\toprule
\textbf{Dimension} & \textbf{Positive} & \textbf{Negative} \\
\midrule
Task    & 222 & 77  \\
Problem & 162 & 33  \\
Method  & 265 & 116 \\
\bottomrule
\end{tabular}
\caption{Dimension-label counts. Counts are over dimension-specific evaluation
instances rather than paper-pair records. The 619 paper-pair records yield 875
pairwise evaluation instances, including 649 positive and 226 negative labels.
The benchmark contains 463 positive paper-pair records, including 236 from ICLR
and 227 from the survey source, and 156 negative paper-pair records. All negative
pairs come from the survey source. Fifty-six negative paper-pair records have
multiple labels.}
\label{tab:dim_counts}
\end{table}

\begin{table}[t]
\centering\small
\setlength{\tabcolsep}{6pt}
\begin{tabular}{lccc}
\toprule
 & \multicolumn{2}{c}{\textbf{Survey}} & \textbf{ICLR} \\
\cmidrule(lr){2-3}\cmidrule(lr){4-4}
\textbf{Dim.} & \textbf{Pos.} & \textbf{Neg.} & \textbf{Pos.} \\
\midrule
Task    & 59.1 & 92.1 & --   \\
Problem & 49.3 & 97.0 & 51.9 \\
Method  & 43.8 & 89.9 & 48.3 \\
\bottomrule
\end{tabular}
\caption{Per-sample accuracy by source, averaged over evaluated models and reported as
percentages. ICLR contributes positive pairs only. Its task dimension is omitted
because only four ICLR task pairs are available.}
\label{tab:source_controlled}
\end{table}

\section{Content Granularity Ablation (Full Results)}
\label{app:granularity}

This section provides the complete content granularity ablation results for all evaluated models across three content modes: \textit{abstract only} (abs), \textit{abstract + introduction} (a+i), and \textit{full paper} (full).

Table~\ref{tab:granularity_ablation} reports macro-averaged metrics over the three dimensions (task, problem, method). The main paper (Table~\ref{tab:main_pairwise}) uses the full-paper setting as default. Figure~\ref{fig:granularity} in the main text highlights five representative models.

\begin{figure*}[t]
\centering
\includegraphics[width=0.72\textwidth]{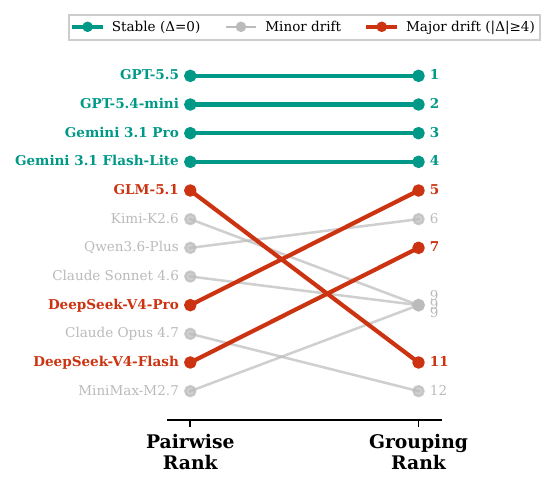}
\caption{Rank drift between pairwise and grouping tasks for the 12 Frontier
models. Top-4 is stable, shown with teal lines, but the mid-pack reorders
substantially, shown with grey lines. GLM-5.1 drops 6 ranks and
DeepSeek-V4-Pro jumps 4 ranks. Red lines highlight major drifts with
$|\Delta| \geq 4$.}
\label{fig:rank_drift}
\end{figure*}

\paragraph{Key observations.}
\begin{itemize}[leftmargin=*,itemsep=2pt]
\item \textbf{F1 generally improves with more content}: Most models show higher prediction accuracy when given full papers compared to abstracts alone.
\item \textbf{Hallucination Rate remains low}: Across all granularities, most frontier models maintain hallucination rates below 2\%, indicating that evidence extraction is generally reliable.
\item \textbf{Mismatch Rate increases substantially with full-paper content}: Models struggle more with nuanced reasoning when given complete papers, with mismatch rates rising from 38--76\% (abstract) to 24--90\% (full paper). This suggests that while models can extract correct evidence from longer texts, they have difficulty maintaining reasoning quality.
\end{itemize}

\section{Rank Drift: Pairwise vs Grouping}
\label{sec:analysis:rank}

The pairwise and grouping tasks described in \S\ref{sec:exp:pairwise} and
\S\ref{sec:exp:multipaper} share the same 12 Frontier models and the same
cascading metric, so their Verified F1 numbers are directly comparable.
Figure~\ref{fig:rank_drift} shows each model's rank on both evaluations.

The top of the leaderboard is stable: GPT-5.5, GPT-5.4-mini, Gemini 3.1
Pro and Flash-Lite hold the top four positions on both evaluations.
The middle of the leaderboard is not: GLM-5.1 drops from 5th to 11th
($-6$ ranks), Claude Opus 4.7 collapses to last place (VF1 = 2.6\%),
while two DeepSeek-V4 variants jump four ranks in the opposite direction.
Multi-paper grouping separates models that sustain grounded reasoning across
documents from those that rely on pairwise shortcuts.


\section{Evidence Judge Validation}
\label{app:judge_validation}

\begin{table}[t]
\centering\small
\setlength{\tabcolsep}{6pt}
{\renewcommand{\arraystretch}{1.12}
\begin{tabular}{@{}lcc@{}}
\toprule
 & \multicolumn{2}{c}{\textbf{Human label}} \\
\cmidrule(l){2-3}
\textbf{Judge label} & \textbf{Unsupported} & \textbf{Supported} \\
\midrule
Unsupported & 41 & 11 \\
Supported   & 2  & 46 \\
\bottomrule
\end{tabular}
}
\caption{Human validation of the LLM-based mismatch judge on 100 stratified
non-hallucinated positive outputs. The judge agrees with human labels in
87.0\% of cases, with Cohen's $\kappa=0.74$.}
\label{tab:human_mismatch_validation}
\end{table}

\begin{table*}[t]
\centering\small
\setlength{\tabcolsep}{10pt}
\begin{tabular}{llccc}
\toprule
\textbf{Model} & \textbf{Dim.} & \textbf{GPT-5.2 self} & \textbf{Gemini-3.1-Pro} & \textbf{$\Delta$} \\
\midrule
GPT-5.5      & task    & 71.6 & 71.5 & $-0.1$ \\
GPT-5.5      & problem & 61.5 & 61.0 & $-0.5$ \\
GPT-5.5      & method  & 63.7 & 43.1 & $\mathbf{-20.6}$ \\
\midrule
GPT-5.4-mini & task    & 53.1 & 51.9 & $-1.2$ \\
GPT-5.4-mini & problem & 36.4 & 37.3 & $+0.9$ \\
GPT-5.4-mini & method  & 36.5 & 29.2 & $\mathbf{-7.3}$ \\
\bottomrule
\end{tabular}
\caption{Verified F1 of GPT-family models under the GPT-5.2 self-judge and the
third-party Gemini-3.1-Pro judge. The larger method gap indicates
self-preference in the default judge.}
\label{tab:judge_validation}
\end{table*}

\subsection{Human Validation of Mismatch Judgments}

To validate the reliability of the LLM mismatch judge, we manually
annotated a stratified sample of 100 non-hallucinated positive outputs.
The sample was balanced across pairwise and grouping tasks, the task, problem, and
method dimensions, and decisions that the judge supported or did not support.
Human annotators judged whether the cited evidence logically supported the
model's stated reason.

Table~\ref{tab:human_mismatch_validation} shows that the LLM judge agrees with
human labels in 87.0\% of cases (95\% Wilson CI: 79.0--92.2), with Cohen's
$\kappa=0.74$.
For identifying unsupported reasoning, the judge achieves 78.8\% precision,
95.3\% recall, and 86.3\% F1.
Only 2 of 48 judge-supported cases were marked unsupported by humans, indicating
that cases passing the judge are rarely false passes under human assessment.
The judge is therefore reliable for detecting unsupported reasoning, while
tending to err conservatively by marking some human-supported cases as
unsupported.

\subsection{Evidence of GPT Self-Preference}

To validate the fairness of our judge routing strategy, we conducted a controlled
experiment with GPT-5.2 as the default judge and Gemini-3.1-Pro as a third-party judge.
We applied both judges to GPT-5.5 and GPT-5.4-mini.

We reran the reasoning check for evidence and reason mismatches with both judges and
report the resulting Verified F1 in Table~\ref{tab:judge_validation}. Hallucination Rate
measures verbatim evidence and does not depend on the judge, so it is unchanged across
the two runs.

\begin{table*}[t]
\centering\small
\setlength{\tabcolsep}{4pt}
\begin{tabular}{lcccccc}
\toprule
 & \multicolumn{3}{c}{\textbf{Raw agreement (\%)}} & \multicolumn{3}{c}{\textbf{Cohen's $\kappa$}} \\
\cmidrule(lr){2-4}\cmidrule(lr){5-7}
\textbf{Judge pair} & \textbf{Abstract} & \textbf{Abstract and intro} & \textbf{Full} & \textbf{Abstract} & \textbf{Abstract and intro} & \textbf{Full} \\
\midrule
GPT-5.2 \& Gemini-3.1-Pro         & 81.2 & 79.6 & 81.1 & 0.57 & 0.54 & 0.52 \\
GPT-5.2 \& DeepSeek-V4-Pro        & 79.2 & 79.3 & 81.2 & 0.55 & 0.55 & 0.53 \\
Gemini-3.1-Pro \& DeepSeek-V4-Pro & 81.9 & 81.7 & 82.5 & 0.62 & 0.62 & 0.60 \\
\bottomrule
\end{tabular}
\caption{Pairwise raw agreement and Cohen's $\kappa$ between the three mismatch judges
for each input setting.}
\label{tab:cross_judge_agreement}
\end{table*}

The results show that on task and problem dimensions, the two judges agree within $\pm 2$ percentage points Verified F1. However, on the method dimension, where evidence is most nuanced, the GPT-5.2 self-judge produces Verified F1 that is 7.3 to 20.6 percentage points higher than the third-party Gemini-3.1-Pro judge. This one-sided inflation is what we mean by \emph{self-preference bias}.

This finding motivates the per-model judge routing used throughout the paper.
GPT-family models are judged by Gemini-3.1-Pro, as shown in Tables~\ref{tab:main_pairwise}
and \ref{tab:main_grouping}. All other models are judged by GPT-5.2.

\subsection{Consistency Across Judges}
\label{app:cross_judge}

To test judge dependence, we reran the mismatch judgment with GPT-5.2,
Gemini-3.1-Pro, and DeepSeek-V4-Pro on Claude Sonnet 4.6 outputs under all three input
settings. Each setting contains 1128--1237 records. Pairwise agreement is 79--83\%
with $\kappa=0.52$--$0.62$. Agreement among all three judges is 66--76\%, with the
lowest value for method, as shown in Tables~\ref{tab:cross_judge_agreement} and
\ref{tab:cross_judge_unanimous}.

On the 60 contested records, GPT-5.2 reaches 80\% accuracy, compared with 55\% for
majority vote, 43\% for Gemini-3.1-Pro, and 32\% for DeepSeek-V4-Pro
as shown in Table~\ref{tab:cross_judge_contested}. Majority voting performs worse
because the two lenient judges make correlated errors. These accuracies on contested
cases are not comparable to the 87.0\% balanced human-validation result above, which
uses a different sample.

\begin{table}[t]
\centering\small
\setlength{\tabcolsep}{6pt}
\begin{tabular}{lccc}
\toprule
\textbf{Dimension} & \textbf{Abstract} & \textbf{Abstract and intro} & \textbf{Full} \\
\midrule
Task    & 73.8 & 72.8 & 75.7 \\
Problem & 72.5 & 72.1 & 73.0 \\
Method  & 66.9 & 66.1 & 67.4 \\
\bottomrule
\end{tabular}
\caption{Percentage of records with the same verdict from all three judges, by dimension
and input setting.}
\label{tab:cross_judge_unanimous}
\end{table}

\begin{table}[t]
\centering\small
\setlength{\tabcolsep}{8pt}
\begin{tabular}{lc}
\toprule
\textbf{Judge or rule} & \textbf{Accuracy} \\
\midrule
GPT-5.2          & 80\% \\
Majority vote    & 55\% \\
Gemini-3.1-Pro   & 43\% \\
DeepSeek-V4-Pro  & 32\% \\
\bottomrule
\end{tabular}
\caption{Accuracy against human labels on the 60 contested records where the judges
disagree.}
\label{tab:cross_judge_contested}
\end{table}

\section{Failed Specialised Models}
\label{app:failed_models}

Two specialised peer-review models could not be evaluated under our structured-output 
protocol:

\textbf{DeepReviewer-7B.} This model disregarded the JSON output format specified 
in the prompt and instead emitted full review-template prose in natural language. 
None of its outputs could be parsed as the required JSON structure containing 
\texttt{is\_similar}, \texttt{evidence\_a}, \texttt{evidence\_b}, and \texttt{reason} 
fields. Manual inspection of 50 outputs confirmed that the model had lost the ability 
to follow structured-output instructions after task-specific supervised fine-tuning 
on peer-review generation.

\textbf{LLaMA-OpenReviewer-8B.} This model produced syntactically valid JSON but 
exhibited severe degeneration in its predictions. In its completed positive-pair
run, covering 649 dimension-specific instances, it predicted
\texttt{is\_similar=false} for 647 of 649 instances and copied the prompt's JSON
schema verbatim into the evidence fields (e.g., \texttt{evidence\_a: "copy-paste
the exact span of text from Paper A..."}). Its available negative-pair run contains
only 60 records and is therefore incomplete. The model appeared to have memorized 
the prompt template during fine-tuning but lost the semantic understanding required 
to instantiate it with actual paper content.

Both failures suggest that task-specific SFT for review generation may erode the 
general instruction-following and structured-output discipline needed for cascading 
evaluation protocols.

\clearpage
\onecolumn
\section{Case Study: Right Label, Wrong Reason}
\label{app:case_study}

The cliff of \S\ref{sec:analysis:cliff} is most legible at the level
of a single pair.
We illustrate with a method-dimension case from Claude~Sonnet~4.6
(full transcript in Appendix~\ref{app:failure_cases}).

\textbf{Diagnosis.}
On the pair \emph{IndicSuperTokenizer} $\leftrightarrow$ \emph{SuperBPE},
the model predicts \texttt{method\,=\,similar} (correct) and cites
verbatim evidence from both papers (Stage~2 passes).
However, Evidence~A states only a high-level contribution (``two-stage
subword--superword learning'') while Evidence~B provides implementation
specifics (whitespace enforcement, lifted restriction, bridging word
boundaries).
The model's reason silently transfers Paper-B specifics onto Paper~A
without textual support, and the cascade correctly flags this as
\textcolor{red!70!black}{mismatched}.

The final label happens to be correct because IndicSuperTokenizer does adopt
the SuperBPE curriculum, but the cascade reveals that the model
reached the right answer through unreliable reasoning.
This is the same contribution-implementation gap identified in
\S\ref{sec:analysis:cliff}, observed in 72.5\% of mismatch cases.
An F1-only metric would miss this failure entirely.

\section{Illustrative Examples}
\label{app:failures}
\label{app:failure_cases}

This section provides two contrasting examples to illustrate the cascading verification pipeline:
\textbf{Case A} demonstrates a verified prediction where all three stages pass,
while \textbf{Case B} shows a mismatched prediction that fails Stage~3 (reasoning verification)
despite passing Stages~1 and~2.

\begin{casebox}{Case A: Verified Prediction (GPT-5.5)}

\noindent\textbf{Paper pair:}\\
\textbf{Paper A:} \emph{Forget Unlearning: Towards True Data-Deletion in Machine Learning}\\
\textbf{Paper B:} \emph{Adaptive Machine Unlearning}

\medskip
\noindent\textbf{Dimension:} method \quad 
\textbf{Gold label:} similar \quad 
\textbf{Prediction:} similar (\checkmark\,correct)

\medskip
\noindent\textbf{Model output (sample 1 of 3):}

\medskip
\noindent\textbf{Evidence A.} 
``We also provide a reduction theorem that if a mechanism is differentially private with respect to the remaining records and satisfies a data-deletion guarantee under non-adaptive edit requests, then it also satisfies a data-deletion guarantee under adaptive requests. Based on this reduction, we redefine the problem of data-deletion as designing a mechanism that (1.) satisfies a data-deletion guarantee against non-adaptive deletion requests, (2.) is differentially private for remaining records, and (3.) has the same utility guarantee as retraining under identical differential privacy constraints.''

\medskip
\noindent\textbf{Evidence B.} 
``We give a generic reduction. We show that if:

\hspace{1em}(1) A data deletion algorithm $R_A$ for a learning algorithm $A$ has deletion guarantees for oblivious sequences of deletion requests (as those from past work do), and

\hspace{1em}(2) Information about the internal randomness of $R_A$ is revealed only in a manner that satisfies differential privacy, then

\noindent $(A, R_A)$ also satisfies data deletion guarantees against an adaptive sequence of deletion requests, that can depend in arbitrary ways on the information that the model provider has made public.''

\medskip
\noindent\textbf{Reason.} 
``Both propose a reduction where a deletion/unlearning guarantee for non-adaptive or oblivious request sequences, together with differential privacy, yields a guarantee for adaptive deletion/update requests.''

\medskip
\noindent\textbf{Cascading verification:}

\medskip
\noindent\textbf{Stage 1 (Correctness):} \textcolor{green!50!black}{$\checkmark$ Pass} --- prediction matches gold label (method-similar)

\noindent\textbf{Stage 2 (Hallucination):} \textcolor{green!50!black}{$\checkmark$ Pass} --- both evidence strings are verbatim substrings of source papers

\noindent\textbf{Stage 3 (Reasoning):} \textcolor{green!50!black}{$\checkmark$ Pass} --- judge critique: ``Both evidence strings describe the same specific reduction mechanism where differential privacy combined with non-adaptive (or oblivious) deletion guarantees implies adaptive deletion guarantees.''

\medskip
\noindent\textbf{Final verdict:} \textcolor{green!50!black}{\textbf{Verified}} --- 
This is the rare case where the model reaches \emph{the right answer for the right reason}. 
All three samples for this pair passed all verification stages.

\end{casebox}

\begin{casebox}{Case B: Mismatched Prediction (Claude Sonnet 4.6)}

\noindent\textbf{Paper pair:}\\
\textbf{Paper A:} \emph{IndicSuperTokenizer: An Optimized Tokenizer for Indic Multilingual LLMs}\\
\textbf{Paper B:} \emph{SuperBPE: Space Travel for Language Models} (Liu et al., 2025)

\medskip
\noindent\textbf{Dimension:} method \quad 
\textbf{Gold label:} similar \quad 
\textbf{Prediction:} similar (\checkmark\,correct)

\medskip
\noindent\textbf{Model output (sample 1 of 3):}

\medskip
\noindent\textbf{Evidence A.} 
``our tokenizer combines linguistically grounded pre-tokenization with a two-stage subword--superword learning process (Liu et al., 2025b), yielding a more compact and semantically faithful vocabulary.''

\medskip
\noindent\textbf{Evidence B.} 
``SuperBPE introduces a simple intervention in the pretokenization step, separating tokenizer training into two discrete phases, wherein the tokenizer (1) first learns subwords (by using pretokenization to prevent merges across whitespace) and then (2) learns superwords (by lifting this restriction).''

\medskip
\noindent\textbf{Reason.} 
``Both papers share the same core mechanism: a two-stage BPE curriculum that first learns subword tokens (with whitespace pretokenization enforced) and then learns superword tokens that bridge word boundaries (by lifting that restriction), producing a vocabulary of both subword and multi-word tokens.''

\medskip
\noindent\textbf{Cascading verification:}

\medskip
\noindent\textbf{Stage 1 (Correctness):} \textcolor{green!50!black}{$\checkmark$ Pass} --- prediction matches gold label (method-similar)

\noindent\textbf{Stage 2 (Hallucination):} \textcolor{green!50!black}{$\checkmark$ Pass} --- both evidence strings are verbatim substrings

\noindent\textbf{Stage 3 (Reasoning):} \textcolor{red!70!black}{$\times$ Fail} --- judge critique: ``Evidence A mentions a `two-stage subword--superword learning process' but does not describe the specific mechanism of enforcing and then lifting whitespace restrictions, nor does it mention BPE, both of which the Reason attributes to Paper A.''

\medskip
\noindent\textbf{Final verdict:} \textcolor{red!70!black}{\textbf{Mismatched}} --- 
Evidence A only states a generic ``two-stage subword--superword learning process'' and cites SuperBPE (Liu et al., 2025b). 
Evidence B provides the specific BPE-curriculum details: whitespace pretokenization enforced in phase 1, then lifted in phase 2 to enable cross-word merges.
The Reason silently transfers all of Paper B's specifics (``whitespace pretokenization enforced,'' ``lifting that restriction,'' ``bridge word boundaries'') onto Paper A without textual support.

\medskip
\noindent\textbf{Failure type:} \emph{Specificity transfer} --- 
The model correctly identifies that Paper A cites Paper B, but assumes the implementation details carry over without verifying that Evidence A actually describes them.
Sample 2 of 3 passed Stage 3 because Evidence B was more detailed, but samples 1 and 3 failed.

\end{casebox}

\section{Cascading Failure Case Studies}
\label{app:cascade_cases}

This appendix provides detailed case studies illustrating failures at each
stage of the cascading verification pipeline: \textbf{Stage~2 (hallucination
detection)} and \textbf{Stage~3 (reasoning mismatch detection)}.
Each case includes the model's output, the source text for verification,
and the judge's determination.

\subsection{Stage 2 Failures: Hallucination}
\label{app:cases:hallucination}

\begin{casebox}{Case H1: Cross-Paper Evidence Contamination (DeepReviewer-14B)}

\noindent\textbf{Paper pair:}\\
\textbf{Paper A:} \emph{Beyond Pass@k: Breadth-Depth Metrics for Reasoning Boundaries}\\
\textbf{Paper B:} \emph{Are Your LLMs Capable of Stable Reasoning?}

\medskip
\noindent\textbf{Dimension:} problem \quad 
\textbf{Gold label:} similar \quad 
\textbf{Prediction:} similar (\checkmark\,correct) \quad
\textbf{Content granularity:} abstract only

\medskip
\noindent\textbf{Model output:}

\medskip
\noindent\textbf{Evidence A} (verbatim):
``To assess the reasoning boundary (the fraction of problems a model can solve) researchers often report Pass@k at large sampling budgets.''

\medskip
\noindent\textbf{Evidence B} (hallucinated):
``To assess the reasoning boundary (the fraction of problems a model can solve) researchers often report Pass@k at large sampling budgets.''

\medskip
\noindent\textbf{Reason:}
``Both papers identify the limitation of using Pass@k at large sampling budgets as a metric for assessing the reasoning boundary of large language models.''

\medskip
\noindent\textbf{Cascading verification:}

\medskip
\noindent\textbf{Stage 1 (Correctness):} \textcolor{green!50!black}{$\checkmark$ Pass} --- prediction matches gold label (similar)

\noindent\textbf{Stage 2 (Hallucination):} \textcolor{red!70!black}{$\times$ Fail} --- Evidence B is \emph{character-for-character identical} to Evidence A (cls\_b: hallucination)

\medskip
\noindent\textbf{Failure type:} \emph{Cross-paper evidence contamination} --- 
The model copied Paper A's text verbatim in place of extracting evidence from Paper B.

\end{casebox}

\begin{casebox}{Case H2: Citation Misattribution as Evidence (DeepSeek-V4 Flash)}

\noindent\textbf{Paper pair:}\\
\textbf{Paper A:} \emph{QWEN Technical Report}\\
\textbf{Paper B:} \emph{MobileLLM: Optimizing Sub-billion Parameter Language Models for On-Device Use Cases}

\medskip
\noindent\textbf{Dimension:} task \quad 
\textbf{Gold label:} not similar \quad 
\textbf{Prediction:} similar (\textcolor{red}{$\times$}\,incorrect) \quad
\textbf{Content granularity:} abstract only

\medskip
\noindent\textbf{Model output:}

\medskip
\noindent\textbf{Evidence A} (hallucinated):
``We present Open Pre-trained Transformers (OPT), a suite of decoder-only pre-trained transformers ranging from 125M to 175B parameters, which we aim to fully and responsibly share with interested researchers.''

\medskip
\noindent\textbf{Evidence B} (hallucinated):
``In this work, we introduce QWEN, the first installment of our large language model series. QWEN is a comprehensive language model series that encompasses distinct models with varying parameter counts.''

\medskip
\noindent\textbf{Cascading verification:}

\medskip
\noindent\textbf{Stage 1 (Correctness):} \textcolor{red!70!black}{$\times$ Fail} --- prediction does not match gold label (predicted similar, actually not similar)

\noindent\textbf{Stage 2 (Hallucination):} \textcolor{red!70!black}{$\times$ Fail} --- Evidence A classified as hallucination (cls\_a: hallucination). Evidence B classified as hallucination (cls\_b: hallucination).

\medskip
\noindent\textbf{Failure type:} \emph{Citation misattribution} --- 
Evidence A describes OPT, which appears in Paper A's reference section, not as the paper's own contribution.

\end{casebox}

\begin{casebox}{Case H3: Fabricated Title as Evidence (DeepSeek-V4 Flash)}

\noindent\textbf{Paper pair:}\\
\textbf{Paper A:} \emph{Certain Generalizations in the Analysis of Variance}\\
\textbf{Paper B:} \emph{A Generalization of Fisher's z Test}

\medskip
\noindent\textbf{Dimension:} task \quad 
\textbf{Gold label:} not similar \quad 
\textbf{Prediction:} similar (\textcolor{red}{$\times$}\,incorrect) \quad
\textbf{Content granularity:} abstract only

\medskip
\noindent\textbf{Model output:}

\medskip
\noindent\textbf{Evidence A} (hallucinated):
``Some New Test Criteria in Multivariate Analysis''

\medskip
\noindent\textbf{Evidence B} (verbatim):
``Wilks (1932) defined a generalized variance and found the appropriate A-criteria for testing certain hypotheses concerning the means, variances, and covariances of k normal multivariate populations from which k independent samples have been drawn.''

\medskip
\noindent\textbf{Reason:}
``Both papers target the task of generalizing classical univariate statistical tests (analysis of variance, Fisher's z test) to multivariate populations, focusing on variance components and hypothesis testing for multiple populations.''

\medskip
\noindent\textbf{Cascading verification:}

\medskip
\noindent\textbf{Stage 1 (Correctness):} \textcolor{red!70!black}{$\times$ Fail} --- prediction does not match gold label (predicted similar, actually not similar)

\noindent\textbf{Stage 2 (Hallucination):} \textcolor{red!70!black}{$\times$ Fail} --- Evidence A classified as hallucination (cls\_a: hallucination). The phrase ``Some New Test Criteria in Multivariate Analysis'' does not appear in Paper A.

\medskip
\noindent\textbf{Failure type:} \emph{Fabricated content} --- 
Evidence A is a plausible-sounding but non-existent phrase.

\end{casebox}

\subsection{Stage 3 Failures: Reasoning Mismatch}
\label{app:cases:mismatch}

\begin{casebox}{Case M1: Evidence Description Error (minimax-m2.7)}

\noindent\textbf{Paper pair:}\\
\textbf{Paper A:} \emph{Iteratively Refined Regularization for Nash Equilibria}\\
\textbf{Paper B:} \emph{Adaptively Perturbed Mirror Descent for Learning in Games}

\medskip
\noindent\textbf{Dimension:} method \quad 
\textbf{Gold label:} similar \quad 
\textbf{Prediction:} similar (\checkmark\,correct) \quad
\textbf{Content mode:} full paper

\medskip
\noindent\textbf{Model output:}

\medskip
\noindent\textbf{Evidence A} (verbatim):
``Instead, we fix the regularization strength at a large value for robustness and achieve convergence by iteratively refining the reference policy.''

\medskip
\noindent\textbf{Evidence B} (verbatim):
``In response, we propose Adaptively Perturbed MD (APMD), which adjusts the magnitude of the perturbation by repeatedly updating the slingshot strategy at a predefined interval.''

\medskip
\noindent\textbf{Reason:}
``Both papers share the fundamental mechanism of keeping regularization/perturbation strength constant while iteratively updating a reference/slingshot strategy to progressively approach Nash equilibrium. Paper A explicitly states this iterative refinement of reference policy with fixed regularization strength, and Paper B describes updating the slingshot strategy at intervals to adjust perturbation magnitude---both represent the same algorithmic backbone of constant-strength regularization with iterative anchor updates in Mirror Descent.''

\medskip
\noindent\textbf{Cascading verification:}

\medskip
\noindent\textbf{Stage 1 (Correctness):} \textcolor{green!50!black}{$\checkmark$ Pass}

\noindent\textbf{Stage 2 (Hallucination):} \textcolor{green!50!black}{$\checkmark$ Pass}

\noindent\textbf{Stage 3 (Reasoning):} \textcolor{red!70!black}{$\times$ Fail} --- supported: false

\medskip
\noindent\textbf{Judge critique:}
``Evidence A mentions fixed regularization strength and iterative refinement of a reference policy, but Evidence B explicitly says the method \emph{adjusts} perturbation magnitude (not keeping it constant) and neither excerpt mentions Nash equilibrium, Mirror Descent, or an algorithmic backbone of constant-strength regularization with anchor updates, so the stated methodological similarity is not supported by the provided text.''

\medskip
\noindent\textbf{Failure type:} \emph{Evidence description error} --- 
The model misrepresents Evidence B by claiming it keeps perturbation strength constant, when it explicitly states the method ``adjusts the magnitude of the perturbation.''

\end{casebox}

\begin{casebox}{Case M2: Reason Exceeds Evidence Support (Claude Opus 4.7)}

\noindent\textbf{Paper pair:}\\
\textbf{Paper A:} \emph{Can Large Language Models Explore In-Context?}\\
\textbf{Paper B:} \emph{Large Language Models as Optimizers}

\medskip
\noindent\textbf{Dimension:} task \quad 
\textbf{Gold label:} similar \quad 
\textbf{Prediction:} similar (\checkmark\,correct) \quad
\textbf{Content mode:} full paper

\medskip
\noindent\textbf{Model output:}

\medskip
\noindent\textbf{Evidence A} (verbatim):
``We evaluate the ability of the current generation of large language models (LLMs) to help a decision-making agent facing an exploration-exploitation tradeoff. We use LLMs to explore and exploit in silos in various (contextual) bandit tasks.''

\medskip
\noindent\textbf{Evidence B} (verbatim):
``We investigate the extent to which contemporary Large Language Models (LLMs) can engage in exploration, a core capability in reinforcement learning and decision making. We focus on native performance of existing LLMs, without training interventions. We deploy LLMs as agents in simple multi-armed bandit environments, specifying the environment description and interaction history entirely in-context, i.e., within the LLM prompt.''

\medskip
\noindent\textbf{Reason:}
``Both papers target the same task of evaluating LLMs' ability to perform exploration/exploitation in (contextual) bandit settings via in-context prompting, deploying LLMs as decision-making agents in MAB environments without specialized training interventions. A researcher would file both under the same sub-topic of LLM-based in-context bandit learning.''

\medskip
\noindent\textbf{Cascading verification:}

\medskip
\noindent\textbf{Stage 1 (Correctness):} \textcolor{green!50!black}{$\checkmark$ Pass}

\noindent\textbf{Stage 2 (Hallucination):} \textcolor{green!50!black}{$\checkmark$ Pass}

\noindent\textbf{Stage 3 (Reasoning):} \textcolor{red!70!black}{$\times$ Fail} --- supported: false

\medskip
\noindent\textbf{Judge critique:}
``While both excerpts clearly study evaluating LLMs on exploration/exploitation behavior in bandit tasks, Evidence A does not state that this is done via in-context prompting or that it excludes training interventions, so the reason over-attributes shared task details not supported by the provided text.''

\medskip
\noindent\textbf{Failure type:} \emph{Reason exceeds evidence support} --- 
The model adds specific details (``via in-context prompting,'' ``without training interventions'') that only appear in Evidence B, incorrectly attributing them to both papers.

\end{casebox}

\begin{casebox}{Case M3: Dimension Confusion (GPT-5.4-mini)}

\noindent\textbf{Paper pair:}\\
\textbf{Paper A:} \emph{Action Robust Reinforcement Learning with Probabilistic Policy Execution Uncertainty}\\
\textbf{Paper B:} \emph{Robust Reinforcement Learning via Adversarial Training}

\medskip
\noindent\textbf{Dimension:} method \quad 
\textbf{Gold label:} similar \quad 
\textbf{Prediction:} similar (\checkmark\,correct) \quad
\textbf{Content mode:} abstract only

\medskip
\noindent\textbf{Model output:}

\medskip
\noindent\textbf{Evidence A} (verbatim):
``we focus on action robust RL with the probabilistic policy execution uncertainty, in which, instead of always carrying out the action specified by the policy, the agent will take the action specified by the policy with probability 1-$\rho$ and an alternative adversarial action with probability $\rho$. We establish the existence of an optimal policy on the action robust MDPs with probabilistic policy execution uncertainty and provide the action robust Bellman optimality equation for its solution.''

\medskip
\noindent\textbf{Evidence B} (verbatim):
``Specifically, we consider two scenarios in which the agent attempts to perform an action $a$, and (i) with probability $\alpha$, an alternative adversarial action $\bar{a}$ is taken, or (ii) an adversary adds a perturbation to the selected action in the case of continuous action space.''

\medskip
\noindent\textbf{Reason:}
``Both papers model action uncertainty by replacing the agent's chosen action with an adversarial alternative with some probability. They each build reinforcement-learning methods on top of that same probabilistic action-corruption mechanism.''

\medskip
\noindent\textbf{Cascading verification:}

\medskip
\noindent\textbf{Stage 1 (Correctness):} \textcolor{green!50!black}{$\checkmark$ Pass}

\noindent\textbf{Stage 2 (Hallucination):} \textcolor{green!50!black}{$\checkmark$ Pass}

\noindent\textbf{Stage 3 (Reasoning):} \textcolor{red!70!black}{$\times$ Fail} --- supported: false

\medskip
\noindent\textbf{Judge critique:}
``The reason identifies a shared problem formulation (probabilistic action corruption) rather than a specific algorithmic mechanism or architectural design, and Evidence B lacks any description of the actual reinforcement learning method used to solve the scenario.''

\medskip
\noindent\textbf{Failure type:} \emph{Dimension confusion} --- 
The reason describes a shared \emph{problem} (action uncertainty), but the pair is labeled as method dimension. The model confuses problem formulation with methodological similarity.

\end{casebox}

\onecolumn
\section{Prompt Templates}
\label{app:prompts}

This section provides the complete prompt templates used for all evaluation tasks in NoveltyBench. We include these prompts to support reproducibility (researchers can replicate our evaluation protocol exactly), transparency (prompt design directly affects model behavior and evaluation outcomes), and extensibility (these templates can be adapted to evaluate novelty in new domains or dimensions).

All prompts follow a consistent structure: (1) a \textbf{Role} definition establishing the evaluator's expertise; (2) a \textbf{Dimension Definition} providing a precise, operational definition of the target dimension; (3) \textbf{Decision Rules} specifying when to mark pairs as similar, with edge-case guidance; (4) an \textbf{Output Format} enforcing structured JSON with evidence extraction and reasoning; and (5) \textbf{Input} with paper content placeholders (\texttt{<TITLE\_A>}, \texttt{<CONTENT\_A>}, etc.) filled at runtime.

We provide seven templates: three for pairwise novelty judgment (task, problem, method), three for multi-paper grouping (task, problem, method), and one for reasoning verification (Stage~3 of the cascading evaluation).

\newtcolorbox{promptbox}[1]{
  colback=gray!3,
  colframe=gray!40,
  fonttitle=\bfseries\small,
  title=#1,
  boxrule=0.4pt,
  breakable,
  top=3mm,
  bottom=3mm,
  left=3mm,
  right=3mm,
  arc=1mm,
  before skip=10pt plus 2pt,
  after skip=10pt plus 2pt
}

\lstset{
  basicstyle=\footnotesize\ttfamily,
  breaklines=true,
  columns=fullflexible,
  keepspaces=true,
  frame=none,
  aboveskip=3pt,
  belowskip=3pt,
  literate={—}{{---}}1 {–}{{--}}1
}

\subsection{Pairwise Novelty Judgment: Task Dimension}

\begin{promptbox}{Pairwise Task Prompt}
\begin{lstlisting}
# Role
You are an expert research analyst. Your task is to judge whether Paper A and Paper B target the same **task**.

# Dimension Definition
`task`: the high-level research objective or application scenario that a paper focuses on.
- A task is specific enough to distinguish research directions (e.g., "open-set object detection", "text-to-video generation", "offline reinforcement learning from human feedback"), yet broad enough that multiple papers can tackle it with different methods.
- The task describes **what** the paper is trying to accomplish in the world, not **how** (method) or **why existing solutions fail** (problem).
- Papers can share a task even if they use entirely different methods; papers can share a method while targeting different tasks.

# Decision Rules
- Mark `is_similar: true` if both papers target the same or closely related specific task setting.
- "Closely related" means a researcher familiar with both would file them under the same sub-topic heading in a literature review.
- Do NOT mark similar merely because both papers belong to the same broad field (e.g., both do NLP, Computer Vision, or Reinforcement Learning).
- Do NOT mark similar based on shared methods or problems alone — focus exclusively on the task setting.
- Use only the provided paper content as evidence; do not invent information.

# Output Format
Output valid JSON only. No markdown, no text outside the JSON object.

When `is_similar` is true:
{
  "is_similar": true,
  "evidence_a": "copy-paste exact span from Paper A identifying its task — verbatim substring, 1–3 sentences",
  "evidence_b": "copy-paste exact span from Paper B identifying its task — same requirements",
  "reason": "1–2 sentences why tasks are same/related — derived STRICTLY from evidence above"
}

When `is_similar` is false:
{
  "is_similar": false,
  "evidence_a": "copy-paste exact span from Paper A identifying its task — verbatim substring",
  "evidence_b": "copy-paste exact span from Paper B identifying its task — verbatim substring"
}

# Input
[PAPER A]
Title: <TITLE_A>
Content type: <CONTENT_TYPE_A>
Content: <CONTENT_A>
[PAPER A END]

[PAPER B]
Title: <TITLE_B>
Content type: <CONTENT_TYPE_B>
Content: <CONTENT_B>
[PAPER B END]
\end{lstlisting}
\end{promptbox}

\subsection{Pairwise Novelty Judgment: Problem Dimension}

\begin{promptbox}{Pairwise Problem Prompt}
\begin{lstlisting}
# Role
You are an expert research analyst. Your task is to judge whether Paper A and Paper B address the same **problem**.

# Dimension Definition
`problem`: the specific technical challenge, bottleneck, limitation, or gap that the paper is trying to solve — the precise reason why existing solutions are insufficient.
- A problem describes **why** current approaches fail or fall short, not what the paper does (method) or what it applies to (task).
- Problems can be shared across entirely different tasks (e.g., "distribution shift at test time" affects both image classification and machine translation).
- A broad goal or application capability (e.g., "improve accuracy", "scale to larger datasets") does NOT constitute a problem.
- The problem must identify a concrete technical obstacle, constraint, or gap that motivates the specific contribution.

# Decision Rules
- Mark `is_similar: true` if both papers address the same specific technical obstacle or bottleneck, even if they operate on different tasks or domains.
- Do NOT mark similar if the papers only share the same task or application domain without a shared bottleneck.
- Do NOT mark similar if the problems are loosely related at a surface level but target clearly different failure modes or constraints.
- Use only the provided paper content as evidence; do not invent information.

# Output Format
Output valid JSON only. No markdown, no text outside the JSON object.

When `is_similar` is true:
{
  "is_similar": true,
  "evidence_a": "copy-paste exact span from Paper A stating its technical problem — verbatim, self-contained, 1–3 sentences",
  "evidence_b": "copy-paste exact span from Paper B stating its technical problem — same requirements",
  "reason": "1–2 sentences why problems are same/related — derived STRICTLY from evidence, no new terms"
}

When `is_similar` is false:
{
  "is_similar": false,
  "evidence_a": "copy-paste exact span from Paper A identifying its technical problem — verbatim, 1–3 sentences",
  "evidence_b": "copy-paste exact span from Paper B identifying its technical problem — verbatim, 1–3 sentences"
}

# Input
[PAPER A]
Title: <TITLE_A>
Content type: <CONTENT_TYPE_A>
Content: <CONTENT_A>
[PAPER A END]

[PAPER B]
Title: <TITLE_B>
Content type: <CONTENT_TYPE_B>
Content: <CONTENT_B>
[PAPER B END]
\end{lstlisting}
\end{promptbox}

\subsection{Pairwise Novelty Judgment: Method Dimension}

\begin{promptbox}{Pairwise Method Prompt}
\begin{lstlisting}
# Role
You are an expert research analyst. Your task is to judge whether Paper A and Paper B propose similar **methods**.

# Dimension Definition
`method`: the core technical approach, algorithmic idea, model architecture, or training/inference strategy that a paper contributes.
- A method describes **how** the paper solves its problem — the specific technical mechanism, not the problem itself or the task.
- Two methods are similar if they share a fundamental algorithmic mechanism or conceptual backbone at a meaningful level of specificity.
- Methods can be shared across different tasks (e.g., contrastive learning, RLHF, knowledge distillation each appear across many application areas).
- A broad technique category (e.g., "uses transformers", "uses diffusion", "uses reinforcement learning") is NOT sufficient — require shared specific mechanisms.

# Decision Rules
- Mark `is_similar: true` if both papers share a fundamental technical mechanism, core algorithmic pipeline, or specific architectural design principle.
- "Shared mechanism" means: if you described the method of one paper to an expert, they would recognize the core idea as the same as the other paper's method.
- Do NOT mark similar based on:
  - Both belonging to the same high-level technique family without sharing a specific mechanism (e.g., "both are diffusion models", "both use attention").
  - Similar problem framing or task setting without shared technical approach.
  - Shared preprocessing, evaluation protocol, or dataset.
  - Both being fine-tuning approaches to LLMs without sharing specific fine-tuning strategy.

# Output Format
Output valid JSON only. No markdown, no text outside the JSON object.

When `is_similar` is true:
{
  "is_similar": true,
  "evidence_a": "copy-paste exact span from Paper A describing its core technical mechanism — verbatim, self-contained, 1–3 sentences",
  "evidence_b": "copy-paste exact span from Paper B describing its core technical mechanism — same requirements",
  "reason": "1–2 sentences stating the specific shared mechanism — derived STRICTLY from evidence, no new terms"
}

When `is_similar` is false:
{
  "is_similar": false,
  "evidence_a": "copy-paste exact span from Paper A identifying its core technical mechanism — verbatim, 1–3 sentences",
  "evidence_b": "copy-paste exact span from Paper B identifying its core technical mechanism — verbatim, 1–3 sentences"
}

# Input
[PAPER A]
Title: <TITLE_A>
Content type: <CONTENT_TYPE_A>
Content: <CONTENT_A>
[PAPER A END]

[PAPER B]
Title: <TITLE_B>
Content type: <CONTENT_TYPE_B>
Content: <CONTENT_B>
[PAPER B END]
\end{lstlisting}
\end{promptbox}

\subsection{Multi-paper Grouping: Task Dimension}

\begin{promptbox}{Grouping Task Prompt}
\begin{lstlisting}
# Role
You are an expert research analyst. You are given <N_PAPERS> research papers. Your task is to identify which papers address the same **task** and group them accordingly.

# Dimension Definition
`task`: the specific application domain, problem setting, or use case that the paper targets — the concrete scenario where the method is applied.
- A task describes **what** the paper is trying to accomplish in terms of input-output behavior or application context, not how it does it (method) or why existing solutions fail (problem).
- Tasks can be shared across different methods (e.g., "image classification", "machine translation", "question answering" each have many different technical approaches).
- A broad research area (e.g., "computer vision", "NLP", "reinforcement learning") is NOT a task — require a specific application setting.
- The task must be concrete enough that a practitioner could recognize whether their use case matches it.

# Grouping Rules
- Place two or more papers in the same group if they target the same or closely related specific task setting.
- "Closely related" means a researcher familiar with both would file them under the same sub-topic heading in a literature review.
- Do NOT group papers merely because they belong to the same broad field (e.g., both do NLP, Computer Vision, or Reinforcement Learning).
- Do NOT group papers based on shared methods or problems alone — focus exclusively on the task setting.
- A paper can belong to at most one group. If a paper does not share a task with any other paper, it does not appear in any group.
- There may be zero, one, or multiple groups.
- Use only the provided paper content as evidence; do not invent information.

# Output Format
Output valid JSON only. No markdown, no text outside the JSON object.

When groups exist:
{
  "groups": [
    {
      "paper_indices": [1, 2],
      "evidence": {
        "1": "copy-paste exact span from Paper 1 identifying its task — verbatim",
        "2": "copy-paste exact span from Paper 2 identifying its task — verbatim"
      },
      "reason": "1–2 sentences why these papers share the same task, referencing evidence above"
    }
  ]
}

When no groups exist:
{"groups": []}

# Input
<PAPERS>
\end{lstlisting}
\end{promptbox}

\subsection{Multi-paper Grouping: Problem Dimension}

\begin{promptbox}{Grouping Problem Prompt}
\begin{lstlisting}
# Role
You are an expert research analyst. You are given <N_PAPERS> research papers. Your task is to identify which papers address the same **problem** and group them accordingly.

# Dimension Definition
`problem`: the specific technical challenge, bottleneck, limitation, or gap that a paper is trying to solve — the precise reason why existing solutions are insufficient.
- A problem describes **why** current approaches fail or fall short, not what the paper does (method) or what it applies to (task).
- Problems can be shared across entirely different tasks (e.g., "distribution shift at test time" affects both image classification and machine translation).
- A broad goal or application capability (e.g., "improve accuracy", "scale to larger datasets") does NOT constitute a problem.
- The problem must identify a concrete technical obstacle, constraint, or gap that motivates the specific contribution.

# Grouping Rules
- Place two or more papers in the same group if they address the same specific technical obstacle or bottleneck, even if they operate on different tasks or domains.
- Do NOT group papers that only share the same task or application domain without a shared bottleneck.
- Do NOT group papers whose problems are loosely related at a surface level but target clearly different failure modes or constraints.
- A paper can belong to at most one group. If a paper does not share a problem with any other paper, it does not appear in any group.
- There may be zero, one, or multiple groups.
- Use only the provided paper content as evidence; do not invent information.

# Output Format
Output valid JSON only. No markdown, no text outside the JSON object.

When groups exist:
{
  "groups": [
    {
      "paper_indices": [1, 2],
      "evidence": {
        "1": "copy-paste exact span from Paper 1 stating its technical problem — verbatim",
        "2": "copy-paste exact span from Paper 2 stating its technical problem — verbatim"
      },
      "reason": "1–2 sentences why these papers address the same bottleneck, referencing evidence"
    }
  ]
}

When no groups exist:
{"groups": []}

# Input
<PAPERS>
\end{lstlisting}
\end{promptbox}

\subsection{Multi-paper Grouping: Method Dimension}

\begin{promptbox}{Grouping Method Prompt}
\begin{lstlisting}
# Role
You are an expert research analyst. You are given <N_PAPERS> research papers. Your task is to identify which papers propose or use the same **method** and group them accordingly.

# Dimension Definition
`method`: the core technical approach, algorithmic idea, model architecture, or training/inference strategy that a paper contributes or relies on.
- A method describes **how** the paper solves its problem — the specific technical mechanism, not the problem itself or the task.
- Two methods are similar if they share a fundamental algorithmic mechanism or conceptual backbone at a meaningful level of specificity.
- Methods can be shared across different tasks (e.g., contrastive learning, RLHF, knowledge distillation each appear across many application areas).
- A broad technique category (e.g., "uses transformers", "uses diffusion", "uses reinforcement learning") is NOT sufficient — require shared specific mechanisms.

# Grouping Rules
- Place two or more papers in the same group if they share a fundamental technical mechanism, core algorithmic pipeline, or specific architectural design principle.
- "Shared mechanism" means: if you described the method of one paper to an expert, they would recognize the core idea as essentially the same as the other paper's method.
- Do NOT group papers based on:
  - Both belonging to the same high-level technique family without sharing a specific mechanism (e.g., "both are diffusion models", "both use attention").
  - Similar problem framing or task setting without a shared technical approach.
  - Shared preprocessing, evaluation protocol, or dataset.
  - Both fine-tuning LLMs without sharing a specific fine-tuning strategy.
- A paper can belong to at most one group. If a paper does not share a method with any other paper, it does not appear in any group.
- There may be zero, one, or multiple groups.
- Use only the provided paper content as evidence; do not invent information.

# Output Format
Output valid JSON only. No markdown, no text outside the JSON object.

When groups exist:
{
  "groups": [
    {
      "paper_indices": [1, 2],
      "evidence": {
        "1": "copy-paste exact span from Paper 1 describing its core technical mechanism — verbatim",
        "2": "copy-paste exact span from Paper 2 describing its core technical mechanism — verbatim"
      },
      "reason": "1–2 sentences explaining the specific shared mechanism, referencing evidence"
    }
  ]
}

When no groups exist:
{"groups": []}

# Input
<PAPERS>
\end{lstlisting}
\end{promptbox}

\subsection{Reasoning Verification (Stage 3)}

The following prompt is used by the external LLM judge (GPT-5.2 or Gemini-3.1-Pro) to determine whether a model's stated reason is logically entailed by its cited evidence strings. This corresponds to Stage~3 of the cascading evaluation protocol.

\begin{promptbox}{Reasoning Verification Prompt}
\begin{lstlisting}
# Role
You are a rigorous logic auditor. You are given two pieces of evidence extracted verbatim from two research papers, plus a conclusion and a reason written by an analyst. Your task is to judge whether the reason is logically supported by the two evidence strings alone.

# What You Are Given
- **Evidence A**: a verbatim excerpt from Paper A.
- **Evidence B**: a verbatim excerpt from Paper B.
- **Dimension**: which aspect is being compared (`task`, `problem`, or `method`).
- **Conclusion**: the analyst's binary judgment (`similar` or `not similar`).
- **Reason**: the analyst's 1-2 sentence explanation of why the conclusion holds.

# Dimension Granularity
The required level of specificity differs by dimension:
- **task**: Both papers must target the same specific application scenario or research objective -- narrow enough that a researcher would file them under the same sub-topic heading. NOT sufficient: same broad field.
- **problem**: Both papers must address the same specific technical bottleneck or failure mode. NOT sufficient: vague shared challenges.
- **method**: Both papers must share a specific algorithmic mechanism or architectural design principle. NOT sufficient: same broad paradigm (e.g., "both use attention").

# Your Task
Judge whether a reader who sees only Evidence A and Evidence B can reasonably arrive at the stated Conclusion via the stated Reason.

Specifically, ask:
1. Does Evidence A actually say what the Reason claims about Paper A?
2. Does Evidence B actually say what the Reason claims about Paper B?
3. Does the Reason correctly characterise the relationship between the two evidence strings?
4. Is the similarity **specific and substantial** at the granularity required for the given Dimension?

# Decision Rules
- Mark `supported: true` if the Reason follows directly from the two evidence strings AND meets the granularity bar.
- Mark `supported: false` if:
  - The Reason attributes content not present in the evidence.
  - The Reason introduces facts beyond what either evidence says.
  - The Reason correctly describes one paper but misrepresents the other.
  - The shared element is too generic for the Dimension.
- Do NOT use your own knowledge about the papers.

# Output Format
Output valid JSON only.
{"supported": true/false, "critique": "One sentence explanation."}

# Input
Dimension: <DIMENSION>
Conclusion: <CONCLUSION>

[Evidence A]
<EVIDENCE_A>
[Evidence A END]

[Evidence B]
<EVIDENCE_B>
[Evidence B END]

Reason: <REASON>
\end{lstlisting}
\end{promptbox}

\end{document}